\documentclass[10pt,twocolumn,letterpaper]{article}

\usepackage{iccv}
\usepackage{times}
\usepackage{epsfig}
\usepackage{graphicx}
\usepackage{amsmath}
\usepackage{amssymb}
\usepackage{booktabs}
\usepackage{comment}
\usepackage{color,colortbl}
\usepackage{xcolor}
\usepackage{xspace}
\usepackage{url}
\usepackage{cite}
\usepackage{multicol}
\usepackage{multirow}
\usepackage{setspace}
\usepackage{tabularx}
\usepackage{diagbox}
\usepackage{caption, subcaption}

\usepackage[pagebackref=true,breaklinks=true,letterpaper=true,colorlinks,bookmarks=false]{hyperref}

\iccvfinalcopy 

\def\iccvPaperID{****} 

\ificcvfinal\fi

\usepackage[capitalize]{cleveref}
\crefname{section}{Sec.}{Secs.}
\Crefname{section}{Section}{Sections}
\Crefname{table}{Table}{Tables}
\crefname{table}{Tab.}{Tabs.}

\newcommand{\disclaimer}{
    \vspace{-1.5cm}
    \begin{center}
        \normalsize\textit{To appear in The 38th Annual AAAI Conference on Artificial Intelligence (AAAI-24)}        
    \end{center}
    \vspace{1.0cm}
}

\begin{document}

\title{\disclaimer NILUT: Conditional Neural Implicit 3D Lookup Tables for Image Enhancement}

\author{Marcos V. Conde$^{1}$, Javier Vazquez-Corral$^{2,3}$, Michael S. Brown$^{4}$, Radu Timofte$^{1}$\\
$^{1}$ \small{Computer Vision Lab, CAIDAS, University of Würzburg} \quad $^{2}$ Computer Vision Center (CVC)\\
 $^{3}$ \small{Department of Computer Science, Universitat Autònoma de Barcelona}
 \quad $^{4}$ York University\\
}

\twocolumn[{
\vspace{-10mm}
\maketitle
\begin{center}
\includegraphics[trim={0 0 0 0},clip, width=\linewidth]{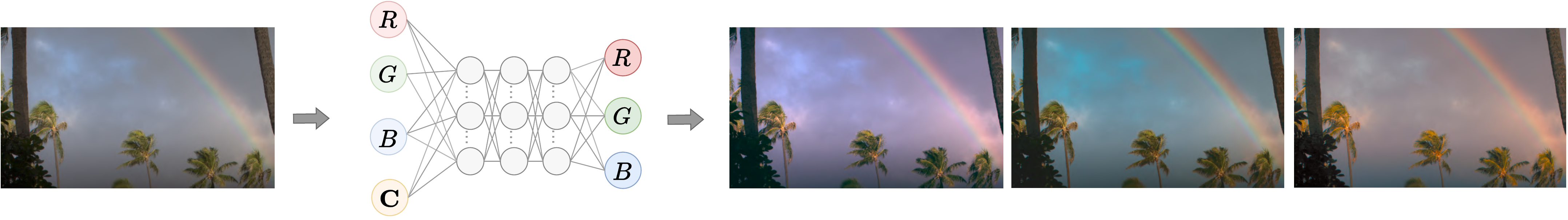}
\put(-348,58){NILUT}
\put(-402,4){Style}
\put(-468,-2){Input $\mathbf{I}$}
\put(-253,-2){\small{Style$=1$ \emph{``Night"}}}
\put(-175,-2){\small{Style$=2$ \emph{``Cyberpunk"}}}
\put(-75,-2) {\small{\emph{Blend} Styles 1\&2}}
\vspace{-0.5mm}
\captionof{figure}{NILUTs encode multiple 3D LUTs in a single representation with the ability to blend between ``styles'' implicitely.}
\label{fig:teaser}
\end{center}
\vspace{1em}
}]

\begin{abstract}
3D lookup tables (3D LUTs) are a key component for image enhancement. Modern image signal processors (ISPs) have dedicated support for these as part of the camera rendering pipeline. Cameras typically provide multiple options for picture styles, where each style is usually obtained by applying a unique handcrafted 3D LUT. Current approaches for learning and applying 3D LUTs are notably fast, yet not so memory-efficient, as storing multiple 3D LUTs is required. For this reason and other implementation limitations, their use on mobile devices is less popular.

In this work, we propose a Neural Implicit LUT (NILUT), an implicitly defined continuous 3D color transformation parameterized by a neural network. We show that NILUTs are capable of accurately emulating real 3D LUTs. Moreover, a NILUT can be extended to incorporate multiple styles into a single network with the ability to blend styles implicitly. Our novel approach is memory-efficient, controllable and can complement previous methods, including learned ISPs. Code, models and dataset available at: \url{https://github.com/mv-lab/nilut}.
\end{abstract}

\section{Introduction}
\label{sec:intro}
 
Image signal processors (ISP) are hardware units used in cameras to process the RAW sensor images to the final output image~\cite{heide2014flexisp,schwartz2018deepisp,karaimer2016software}. The ISP hardware applies a series of processing steps to render the RAW image to its final photo-finished output. 3D lookup tables (\textbf{3D LUTs}) are one of the core components used in conventional ISPs. Specifically, a 3D LUT is a global color operator that maps an RGB color to a new RGB color. 3D LUTs are commonly used to model a desired stylistic look as shown in Fig.~\ref{fig:teaser}. Most cameras can render the same image to several different pictures, where each picture style has its own associated 3D LUT to enhance the color and tone~\cite{zeng2020lut, delbracio2021mobile, zhao2022learning, karaimer2018improving}.

Modern cameras use conventional ISPs~\cite{delbracio2021mobile, karaimer2016software, conde2022model} and apply further photo-finishing deep-learning models~\cite{ignatov2020replacing, tseng2022neural}. These usually run on dedicated processors \eg neural processing units (NPUs), with important limitations such as memory allocation and allowed operations~\cite{ignatov2018ai}.
In addition, there is already an active trend to replace many conventional ISP components with deep learning-based algorithms, \eg, image denoising~\cite{abdelhamed2018high}, color constancy~\cite{Hernandez-Juarez_2020_CVPR}, super-resolution~\cite{agustsson2017ntire, ignatov2021realnpu}, and even the entire ISP~\cite{DeepFlexISP, liang2021cameranet, liu2022deep}.

In particular, methods for image enhancement via color and tone manipulation are often based on 3D LUTs~\cite{zeng2020lut, yang2022adaint} due to their runtime efficiency. However, many of these methods are not suitable for mobile devices due to their memory limitations (\ie storing multiple 3D LUTs would be too memory exhaustive) and required operations.

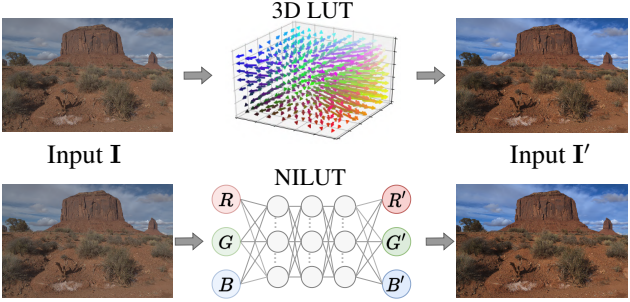
\begin{figure}[t]
  \centering
   \includegraphics[width=\linewidth]{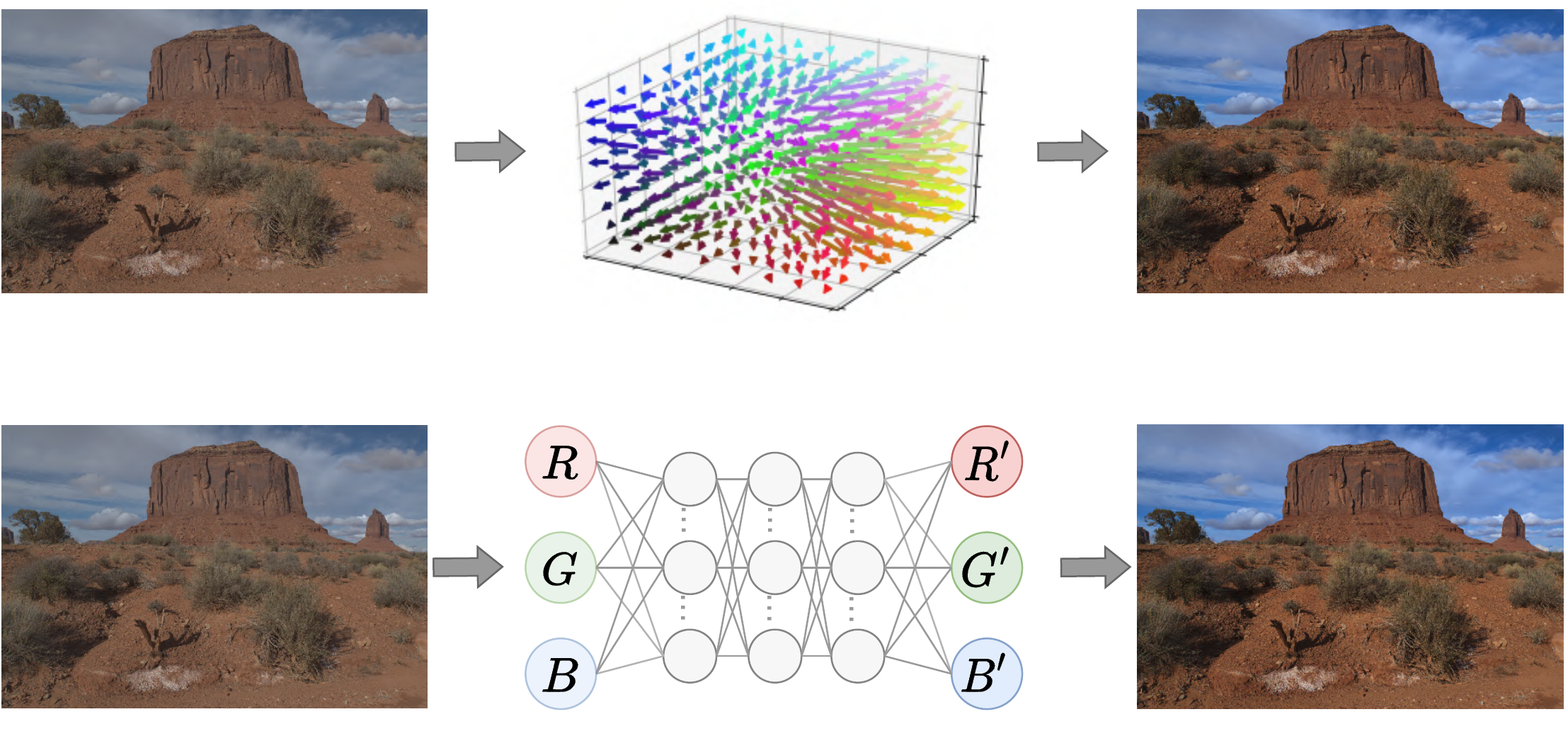}
   \put(-220,55){Input $\mathbf{I}$}
   \put(-45,55) {Input $\mathbf{I'}$}
   \put(-134,47){\small{NILUT}}
   \put(-135,110){\small{3D LUT}}
   \caption{\textbf{Top:} shows a conventional 3D lookup table able to enhance the color and tone of the image. \textbf{Bottom:} shows the same functionality based on the proposed NILUT.}
   \label{fig:teaser2}
\end{figure}

\vspace{-4mm}
\paragraph{Contribution} We propose \textbf{NILUT}, a novel application of implicit neural representations (\textbf{INRs})~\cite{sitzmann2020siren} for color manipulation.
Our NILUT is an implicitly defined, continuous 3D color transformation parameterized by a neural network. NILUTs can accurately mimic existing professional 3D LUTs, as shown in Fig.~\ref{fig:teaser2}.
Moreover, NILUTs can be extended to encode multiple styles into a single network. During inference, the NILUT can be conditioned on a particular picture style, and even blend between multiple styles implicitly. 
This novel multi-style formulation allows controllable image enhancement and customization.
We believe NILUTs can complement previous methods for image enhancement~\cite{zeng2020lut, yang2022adaint, wang2021real, zhao2022learning}, including learned ISPs~\cite{ignatov2020replacing, ignatov2021learnednpu} designed for smartphones.
As part of this effort, we provide a dataset of curated 3D LUTs and images for evaluation.

\section{Related Work}
\label{sec:rel_work}

\subsection{3D LUTs for Color Manipulation}
\label{sec:rel-luts}

3D LUTs are a mechanism to approximate a nonlinear 3D color transform by sparsely sampling the transformation by a discrete 3D lattice~\cite{karaimer2016software, lin2012nonuniform, zeng2020lut}. The model is defined as a mapping $\phi$, usually in the RGB color space, where an input color $\mathbf{I}=[r,g,b]$ is mapped into $\mathbf{I'}=[r',g',b']$:
\begin{equation}
    \phi : \mathbb{R}^{3} \mapsto \mathbb{R}^{3} \quad \phi(\mathbf{I}) = \mathbf{I'}.
\label{eq:3dlut}
\end{equation}

3D LUTs appear at different stages of the image signal processor pipeline, as detailed in \cite{karaimer2016software, karaimer2018improving}. 3D LUTs are often manually created by camera engineers and professional photographers. Methods such as lattice regression~\cite{garcia2009lattice, lin2012nonuniform} parameterize the 3D LUT's lattice from the sparse samples using various regularizations.

More recently, deep learning techniques have been employed for estimating 3D LUTs and learning interpolation and sampling strategies within the LUT with the goal of color image enhancement~\cite{liu20224d,Yang_ECCV,wang2021real}. For example, Zeng~\etal~\cite{zeng2020lut} proposed a method that learned the parameters of three 3D LUTs and an additional per-image adaption to blend between the LUTs. Similarly, Wang~\etal~\cite{wang2021real} proposed a similar idea to learn the parameters of 3D LUTs, but included a spatial blending over the image, effectively implementing a spatially varying 3D LUT. Yang~\etal~AdaInt\cite{yang2022adaint} focused on improving the sampling within the 3D LUTs (based on~\cite{zeng2020lut}). They proposed a learned method to improve the classical trilinear interpolation between uniform sampling points in the LUT~\cite{lin2012nonuniform}. Also, Yang~\etal~\cite{Yang_ECCV} proposed to learn separated component-correlated sub-transforms as 1D and 3D LUTs.

The aforementioned methods focus on conventional 3D LUTs and their interpolation mechanisms.  In contrast, our NILUT is interested in providing the functionality of the 3D LUTs, but as a neural network to be more compatible with NPU-based hardware. In addition, we are interested in encoding multiple picture styles within the same network.

\subsection{ISPs and Image Enhancement}

In recent years, most low-level computer vision tasks have witnessed promising results from deep learning methods. For example, significant advances in image denoising~\cite{abdelhamed2018high, zhang2017beyond}, image deblurring~\cite{nah2021ntire}, image super-resolution~\cite{agustsson2017ntire, liang2021swinir}, and image enhancement \cite{gharbi2017deep,zhang2021star, tu2022maxim, moran2020deeplpf, guo2020zero,wang2020experiment, conde2023perceptual, zhao2022learning, Ma_2022_CVPR} amongst many other tasks. Most of these tasks are crucial stages in modern smartphone ISPs.

Moreover, recent end-to-end learned ISPs~\cite{ignatov2020replacing,liang2021cameranet,liu2022deep, DeepFlexISP} have also obtained promising results.
Due to this current deep learning trend, smartphone manufacturers are incorporating special neural processing units (NPUs)~\cite{ignatov2018ai, web1,web2,web3,web4,web5,web6}.

This said, the previously introduced methods are designed to run on consumer GPUs (\eg NVIDIA V100), and therefore integrating such powerful tools into a smartphone is extremely challenging, and sometimes impossible due to the memory and computational limitations~\cite{ignatov2018ai}.

Our NILUTs represent a new plug-and-play module for modern deep learning-based ISPs and image enhancement pipelines such as Ignatov~\etal real-time image super-resolution~\cite{ignatov2021realnpu} and end-to-end learned ISPs~\cite{ignatov2021learnednpu} tested in real commercial smartphones NPUs. 

\subsection{Implicit Neural Representations}
\label{sec:inrs}

In recent years, implicit neural representations (INRs)~\cite{sitzmann2020siren, genova2019learning, muller2022instant} have become increasingly popular in image processing as a novel way to parameterize an image. Also known as coordinate-based networks, these approaches use multilayer perceptrons (MLPs) to overfit to the image.  
Multiple works have demonstrated the potential of MLPs as continuous, memory-efficient implicit representations for images~\cite{sitzmann2020siren, strumpler2022implicit}; we find especially inspiring SIREN~\cite{sitzmann2020siren} and Fourier Feature Networks~\cite{tancik2020fourier}.
This technique was also successfully applied to model shapes~\cite{genova2019learning, michalkiewicz2019implicit} and 3D scenes~\cite{mildenhall2021nerf, muller2022instant}.

Conventional signal representations are usually discrete \eg, an image is a discrete grid of pixels (subspace of $\mathbb{R}^{2}$) with output values bounded in an $\mathbb{R}^{3}$ RGB space. 
In contrast, INRs parameterize a signal as a continuous function that maps the source domain $\mathcal{S}$ of the signal (\ie, a coordinate) to its corresponding value in the target $\mathcal{T}$ (\ie, the corresponding RGB intensity value). 
This function is approximated using a neural network, and therefore it is not analytically tractable. This can be formulated as:
\begin{equation}
    \Phi : \mathbb{R}^{2} \mapsto \mathbb{R}^{3} \quad \mathbf{x} \to \Phi(\mathbf{x})=[r,g,b],
\label{eq:inr}
\end{equation}
where $\Phi$ is the learned INR function, the domains $\mathcal{S} \in \mathbb{R}^{2} $ and $\mathcal{T} \in \mathbb{R}^{3}$, the input coordinates $\mathcal{\mathbf{x}}$, and the output RGB value $[r,g,b]$. Note that the behavior of this function is similar to a lookup table---see \eqref{eq:3dlut}--- yet being continuous, differentiable, and learnable. Our new application of INRs consists of learning a mapping between two  color representations.

\section{Neural Implicit LUT}
\label{sec:ours}

As discussed in the prior section, the most common INRs in the literature~\cite{sitzmann2020siren, tancik2020fourier} are coordinate-based representations of image signals, implemented using MLPs. 
In this work, our goal is to learn a 3D transformation in the RGB color space, therefore we map $\mathbb{R}^{3}$ coordinates (\ie{} color values $\mathbf{I}$) from a source $\mathcal{S}$ to a target $\mathcal{T}$, being both 3D domains.

Specifically, we want a continuous function $\Phi$ with the following properties:
\begin{align} 
\Phi : \mathbb{R}^{3} \mapsto \mathbb{R}^{3} \quad \Phi(\mathbf{I}) \in [0,1]^3, \\
\Phi (\mathbf{I}) \approx \phi (\mathbf{I}), \\
\nabla\Phi \approx \nabla\phi.
\end{align}

We represent the RGB space as a set $\mathcal{X}=\{x_{i}\}$ of color pixels $x_i=(r_i, g_i, b_i)$. This set contains $\approx16$ million elements if we consider the complete RGB space (\ie{ $256^3$}).
To learn our continual representations $\Phi$ we minimize:

\begin{equation}\label{minimization}
\mathcal{L} = \sum_{i} \| \Phi(x_i) - \phi(x_i)\|_1,
\end{equation}
where $\phi$ is the real 3D LUT -see Eq. \ref{eq:3dlut}-.

This function $\Phi$ is an implicit neural representation of a 3D lookup table $\phi$ and can be formulated as: 

\begin{equation} \label{eq:mlp}
\begin{split}
\Phi (x) = \mathbf{W}_n ( \varsigma_{n-1} \circ \varsigma_{n-2} \circ \ldots \circ \varsigma_0 )(x) + \mathbf{b}_n \\
\varsigma_i (x_i) = \alpha \left( \mathbf{W}_i \mathbf{x}_i + \mathbf{b}_i \right),
\end{split}
\end{equation}
where $\varsigma_i$ are the layers of the network (considering their corresponding weight matrix $\mathbf{W}$ and bias $\mathbf{b}$), and $\alpha$ is a nonlinear activation. We study three different networks: (i) simple ReLU-MLPs or Tanh-MLPs~\cite{sitzmann2020siren}, (ii) SIREN~\cite{sitzmann2020siren} and (iii) Residual MLPs (referred as MLP-Res).

We note here that SIREN~\cite{sitzmann2020siren} has some drawbacks in terms of its implementation. First, it can not use INRs speed-up techniques, such as the one in \cite{muller2022instant}, and second, their custom activations are still not supported for mobile devices accelerator hardware.

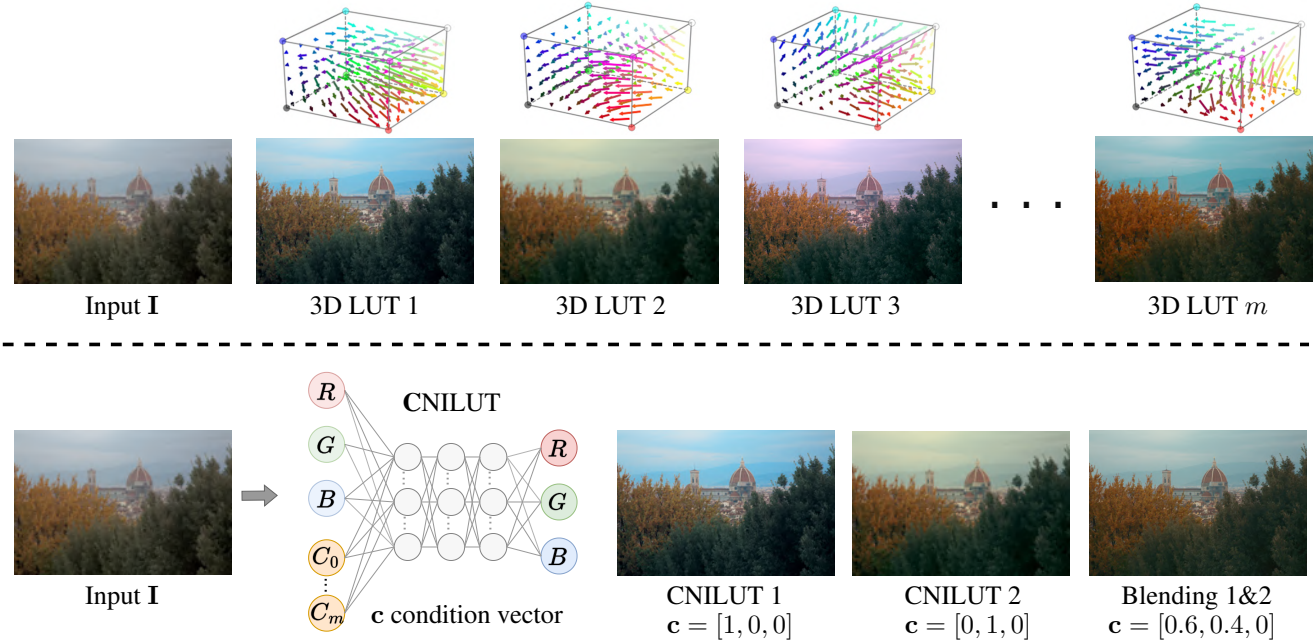
\begin{figure*}[!ht]
  \centering
   \includegraphics[width=\linewidth]{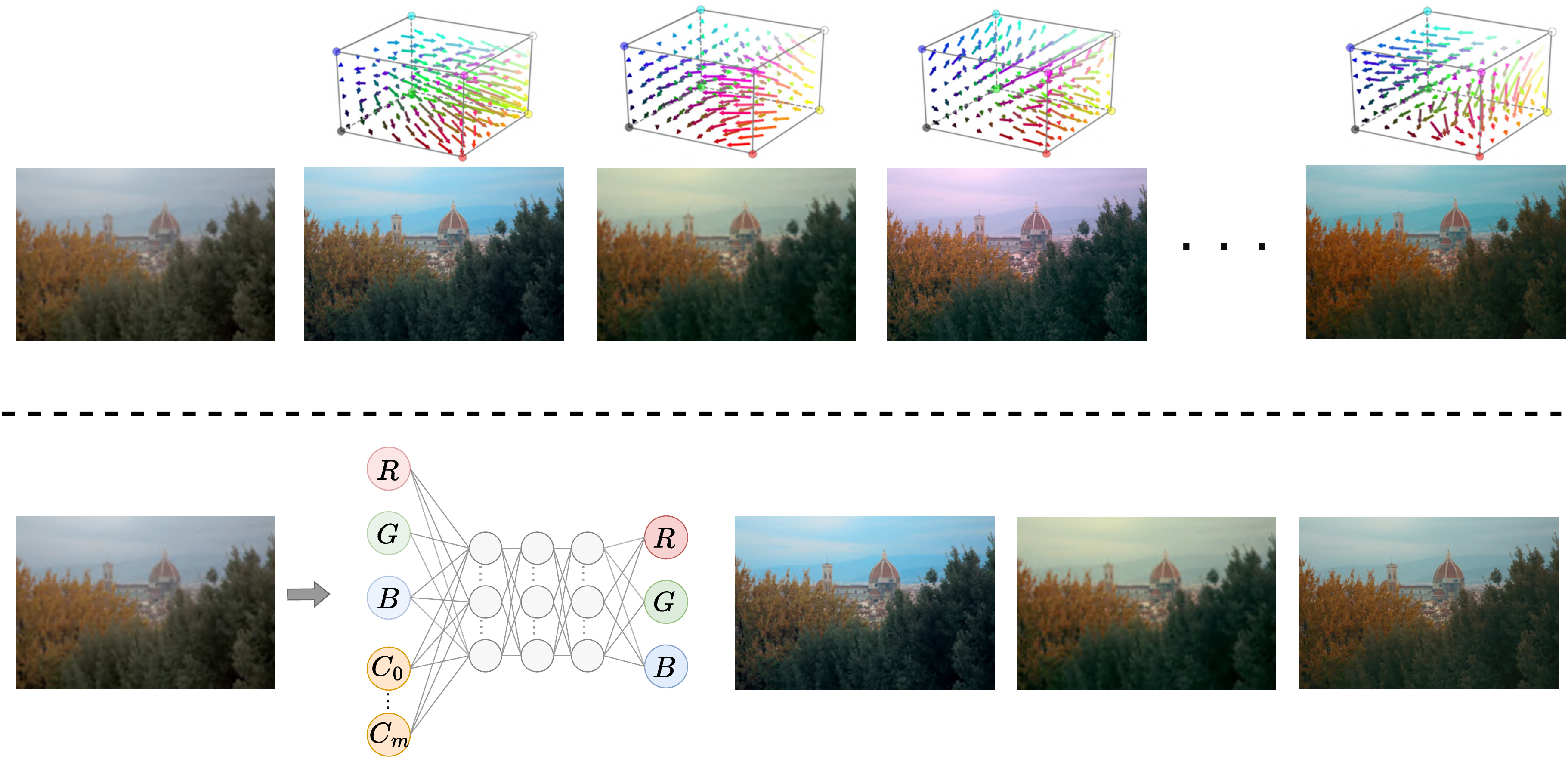}
   \put(-465,12) {Input $\mathbf{I}$}
   \put(-345,85){\textbf{C}NILUT}
   \put(-357,5){$\mathbf{c}$ condition vector}
   \put(-245,12) {CNILUT 1}
   \put(-245,1) {$\mathbf{c}=[1,0,0]$}
   \put(-155,12) {CNILUT 2}
   \put(-155,1) {$\mathbf{c}=[0,1,0]$}
   \put(-73,12) {Blending 1\&2}
   \put(-78,1) {$\mathbf{c}=[0.6,0.4,0]$}
   \put(-465,122){Input $\mathbf{I}$}
   \put(-380,122){3D LUT 1}
   \put(-287,122){3D LUT 2}
   \put(-198,122){3D LUT 3}
   \put(-63,122){3D LUT $m$}
   \caption{\textbf{Top:} A conventional 3D LUT framework. Each 3D LUT is stored and processed individually. \textbf{Bottom:} The new framework introduced by CNILUTs. We use as input both the image and a condition vector (one-hot encoding of the style), allowing for \textbf{i)} selection of multiple styles with a single network, and \textbf{ii)} blending different 3D LUTs styles by modifying the input condition vector. This happens implicitly without additional computational cost.}
   \label{conditional}
\end{figure*}

Another alternative to SIREN would be a residual-based MLP (MLP-Res).  This approach assumes the 3D LUT does not make drastic color changes (e.g., red becoming blue), but instead performs color manipulation via reasonable displacements (residual) between input and output RGB values. Our visualizations of 3D LUTs in Fig.~\ref{conditional} help to reveal that this is often the case. This approach also serves to help regularize the MLP in regions with no changes (i.e., the residual is 0 over the three changes). Our ablations in Section~\ref{sec:exp} also reveal this is a good strategy.

\paragraph{Conditional Neural LUT}
We can further improve the proposed representation to model more complex relationships in the image color domain:
\begin{equation}
    \Psi : \mathbb{R}^{3+m} \mapsto \mathbb{R}^{3} \quad \Psi(\mathbf{z}) = \mathbf{I}',
\end{equation}
where $\mathbf{z} = [\mathbf{I}, \mathbf{c}]$ is the concatenation of the input RGB intensity $\mathbf{I} \in \mathbb{R}^{3}$, and a condition vector $\mathbf{c}\in\mathbb{R}^{m}$ with $m$ possible styles or LUTs using one-hot class encoding. Therefore this continual function $\Psi$ maps an input intensity $\mathbf{I}$ into $\mathbf{I}'$ under the condition $\mathbf{c}$, representing a conditional neural implicit LUT, a more general and powerful representation than the previously introduced NILUT ($\Phi$). 
We illustrate this in Fig.~\ref{conditional}, where we show the codification of the style as a condition vector using one-hot encoding. The derivation of our CNILUT also allows us to blend among different styles by just modifying the values of the condition vector. Details of this are provided in Section~\ref{ssc:cnilut}.

\subsection{Discussion}
\label{sec:discussion}

We end this section by discussing some benefits and differences between NILUTs over conventional 3D LUTs.

Firstly, previous approaches for learning 3D LUTs~\cite{garcia2009lattice, zeng2020lut, yang2022adaint} are notably fast and faithful, allowing real-time processing using regular GPUs (\eg 24Gb memory). We understand that a neural network (\eg NILUT) is limited and cannot surpass the efficiency of a lookup operation. However, in the mobile devices scenario, 3D LUTs have two important limitations:
\textbf{(i)} allocating in memory multiple 3D LUTs (usually 33-dim, thus $107K$ FP parameters~\cite{zeng2020lut}) is not possible due to the hard memory limitations found in mobile devices chips \eg NPUs~\cite{ignatov2018ai}. Note that dedicated processors such as the ISP usually have their own memory and typically only compact models can run on these~\cite{ignatov2021realnpu, ignatov2021learnednpu}.
\textbf{(ii)} many operations, including Zeng~\etal trilinear interpolation on CUDA~\cite{zeng2020lut} are not supported by PyTorch Mobile or TFLite, the most common frameworks for developing efficient mobile models.
We believe these are the reasons why there are very few works about image enhancement on mobile devices using 3D LUTs. Thus, we aim to complement previous approaches and offer a memory-efficient alternative for mobile devices.

Secondly, NILUTs offer other benefits such as being naturally differentiable, allowing end-to-end learning, and are mobile-ready allowing to complement deep learning-based image processing pipelines as a plug-and-play module, in mobile devices.
Finally, NILUTs, as a novel representation, are conditional, allowing a single compact neural network to deal with multiple styles or LUTS that are presented in the imaging process. This clearly contrasts with current LUTS, in which a different process should be run for each specific one. This is key to being memory-efficient as one NILUT can mimic the transformation of five real 3D LUTs. Also, this property allows us to blend among the different styles at inference time in contrast with current LUTs where each blending is individually computed either at the 3D LUT or at the image level (\eg as in~\cite{zeng2020lut}).

\section{Experiments}
\label{sec:exp}

\begin{figure}[!t]
  \centering
   \includegraphics[width=0.92\linewidth]{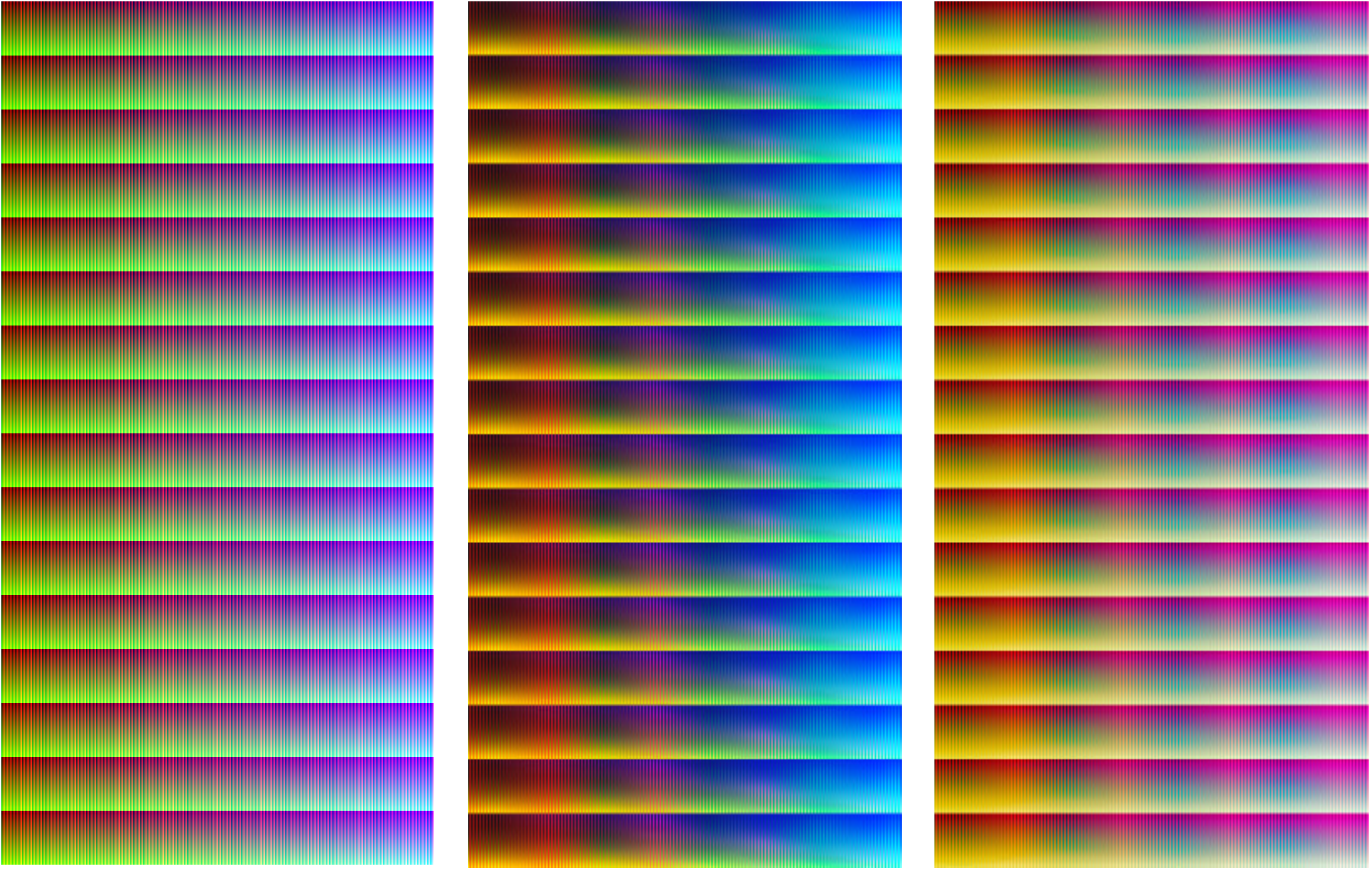}
   \caption{From left to right, original RGB map, output of 3D LUT ``Cyberpunk", output of 3D LUT style ``Nightcolors". We can appreciate the color transformation clearly.
   In graphics, this is referred to as a \textbf{Hald} image, a graphical representation of 3D LUT in the form of a color table that contains all of the color gradations of 3D LUT~\cite{web-3dlut-creation}.
   \vspace{-2mm}
   }
   \label{fig:rgb_map}
\end{figure}

Learning a complete 3D LUT transformation of the 8-bit RGB space requires $256^3=16.78$ million input (and output) colors. According to Eq.~\eqref{minimization}, we represent these $16M$ colors of the RGB space as a set $\mathcal{X}=\{x_i\}$ of size $16M \times 3$. This set is equivalent to an image of dimension $4096\times 4096 \times 3$ that we call $\mathcal{M}$. We will denote this image as the RGB map, illustrated in Fig.~\ref{fig:rgb_map}.  
We then process image $\mathcal{M}$  using professional image editing software (Adobe Photoshop) and real 3D LUTs designed by photographers. These processed images are reshaped back to dimension $16M \times 3$, and correspond to $\phi (x_i)$ on Eq.~\eqref{minimization}, \ie our ground-truth for the minimization.

The coordinate-based MLP $\Phi$, as it is the standard practice with INRs~\cite{sitzmann2020siren}, is trained to ``overfit'' the mapping between its result to the input colors ($\Phi (x_i)$)  and the output of the real LUT  $\phi (x_i)$, as previously introduced in Section \ref{sec:inrs}. We provide visualizations of this training for a subset of the colors in $\mathcal{M}$ in the supplementary material.

Note that using this setup we do not require natural images to learn real 3D LUTs, just the corresponding RGB maps (Halds). This is indeed the way professional photographers create 3D LUTs~\cite{web-3dlut-creation}.

\begin{table}[t]
    \centering
    \resizebox{\linewidth}{!}{
    \begin{tabular}{c c c c c c c}
        \toprule
         Method & N & L & PSNR$_\mathbf{\textbf{rgb}}$ $\uparrow$ & $\Delta$E$_{\textbf{rgb}}$ $\downarrow$ & PSNR$_\mathbf{\textbf{5k}}$ $\uparrow$ & $\Delta$E$_{\textbf{5k}}$ $\downarrow$ \\
         \midrule
         SIREN~\cite{sitzmann2020siren } & 128 & 2 & 44.43 & 1.04 & \underline{41.17} & \textbf{1.63} \\
         SIREN~\cite{sitzmann2020siren } & 64 &  2 & \textbf{45.37} & \textbf{0.96} & 40.35 & 1.82 \\
         MLP-Res & 128 & 2 & \underline{45.34} & \underline{0.97} & \textbf{42.04} & \underline{1.65} \\
         MLP-Res & 64  & 2 & 43.84 & 1.11 & 40.34 & 1.93 \\
         MLP & 128 & 2 & 43.18 & 1.19 & 39.70 & 2.20 \\
         MLP & 64  & 2 & 41.41 & 1.41 & 38.34 & 2.36 \\
         \bottomrule
    \end{tabular}}
    \caption{Evaluation of different NILUT architectures on both the RGB map image and on the MIT5K dataset~\cite{fivek}. We refer to the number of neurons as (N), number of layers as (L). Compact SIREN~\cite{sitzmann2020siren} and MLP-Res architectures can model complex 3D LUTs and approximate their transformations. Best results in bold. Second best underscored.
    \vspace{-3mm}
    }
    \label{tab:ablation2}
\end{table}

\vspace{-4mm}
\paragraph{Evaluation}
We consider two different metrics for our work. We report PSNR, and CIELAB $\Delta$E error. We choose these two measures because i) PSNR is a standard fidelity metric in the literature and ii) $\Delta$E is a perceptual color difference metric that measures differences between two colors \cite{sharma2017digital}, and therefore well suited for our problem. 
Considering that we fit our NILUTs using RGB maps as mentioned before, we evaluate the quality of our NILUT representation in two different ways: (i) RGB mapping quality (Fig.~\ref{fig:rgb_map}), (ii) using natural unseen images from MIT5K~\cite{fivek}.

Our dataset consists on a set of 5 professional 3D LUTs. We report the average metrics over the five.

\textbf{I)} We compare the fidelity between the NILUT and real 3D LUT RGB maps. More in detail, we evaluate the differences between the NILUT-generated map $\Phi (\mathcal{M})$ and the real 3D LUT output map $\phi (\mathcal{M})$ -see Eq.\eqref{minimization}-. These results are referred as PSNR$_\mathbf{\textbf{rgb}}$ and $\Delta$E$_{\textbf{rgb}}$ in Table~\ref{tab:ablation2}

\vspace{1mm}

\textbf{II)} We randomly selected 100 RAW images from the Adobe MIT5K dataset~\cite{fivek}, captured using diverse DSLR cameras. We then processed the images using Adobe Photoshop and the same set of 3D LUTs discussed before. Once a NILUT is fitted using the corresponding RGB map, we apply it to this set of images and measure the fidelity between the real 3D LUT and NILUT processed images. These results are referred as PSNR$_\mathbf{\textbf{5k}}$ and $\Delta$E$_{\textbf{5k}}$ in Table~\ref{tab:ablation2}.

\vspace{2mm}

Table~\ref{tab:ablation2} presents our results for different configurations of MLPs -a basic MLP, SIREN, and a Residual MLP-, under two different numbers of neurons (N) and layers (L). Note that differences in $\Delta$E  smaller than 2 are indistinguishable by human observers~\cite{sharma2017digital}. 
Also, results with more than 40 dB PSNR are considered of high quality.
Attending to Table~\ref{tab:ablation2} results, we can affirm that NILUTs can mimic almost perfectly the RGB transformation of real 3D LUTs on natural images from photographers~\cite{fivek}.

\begin{figure}[!ht]
    \centering
    \setlength\tabcolsep{1pt}
    \begin{tabular}{ccc}
    \includegraphics[width=0.30\linewidth]{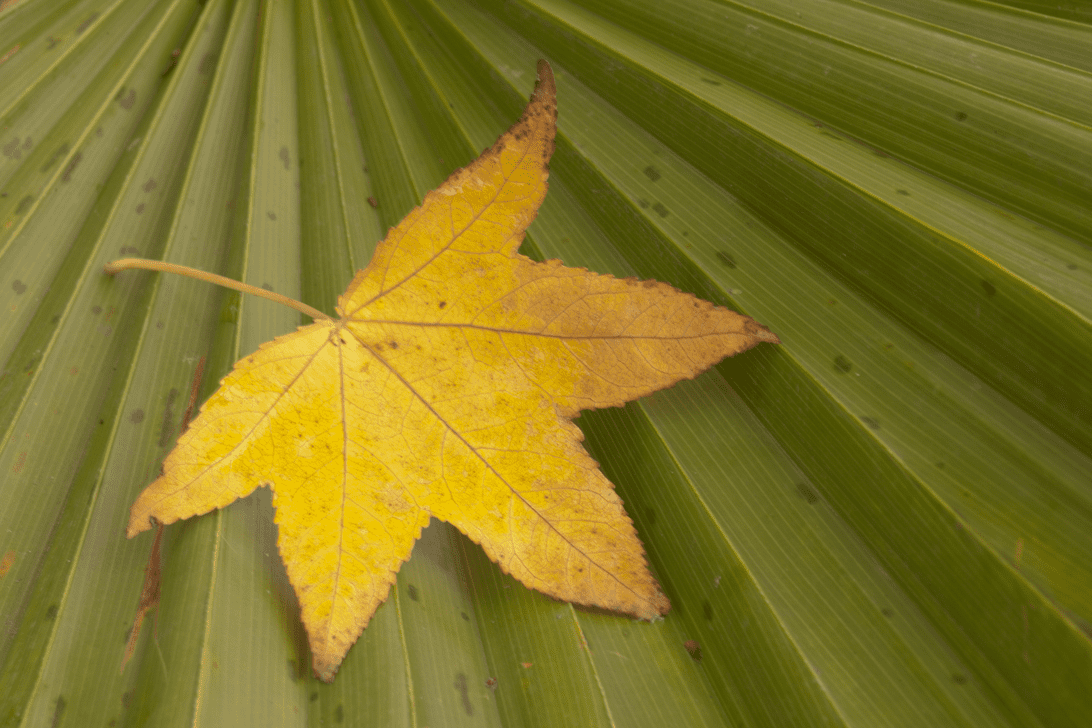} &
    \includegraphics[width=0.30\linewidth]{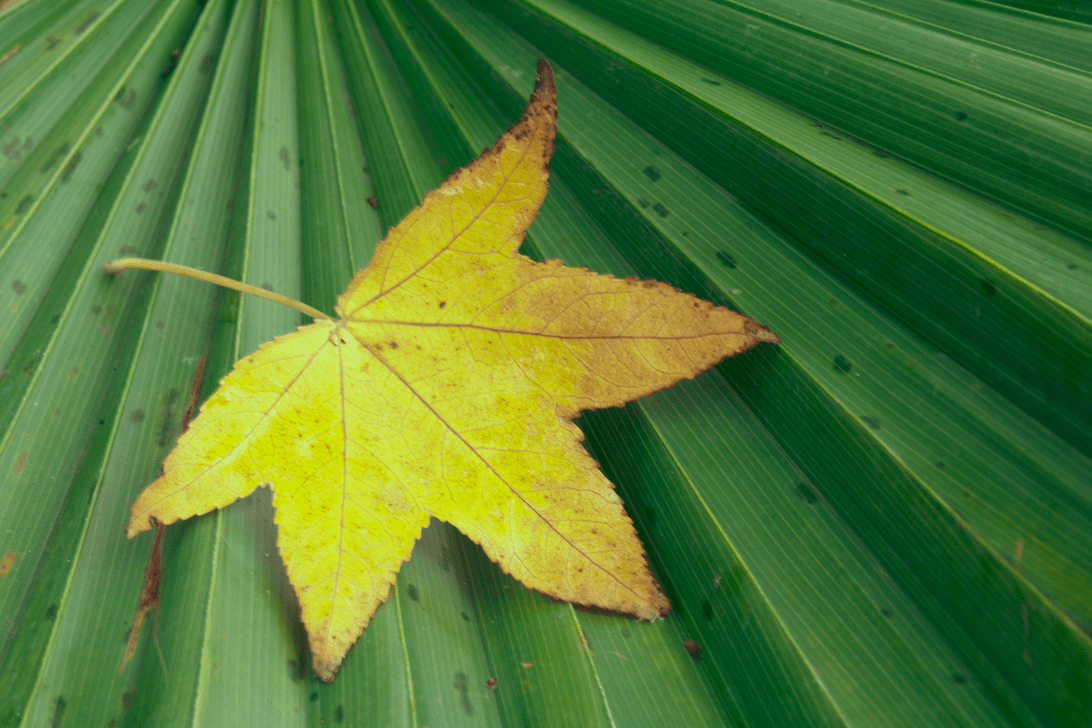} &
    \includegraphics[width=0.30\linewidth]{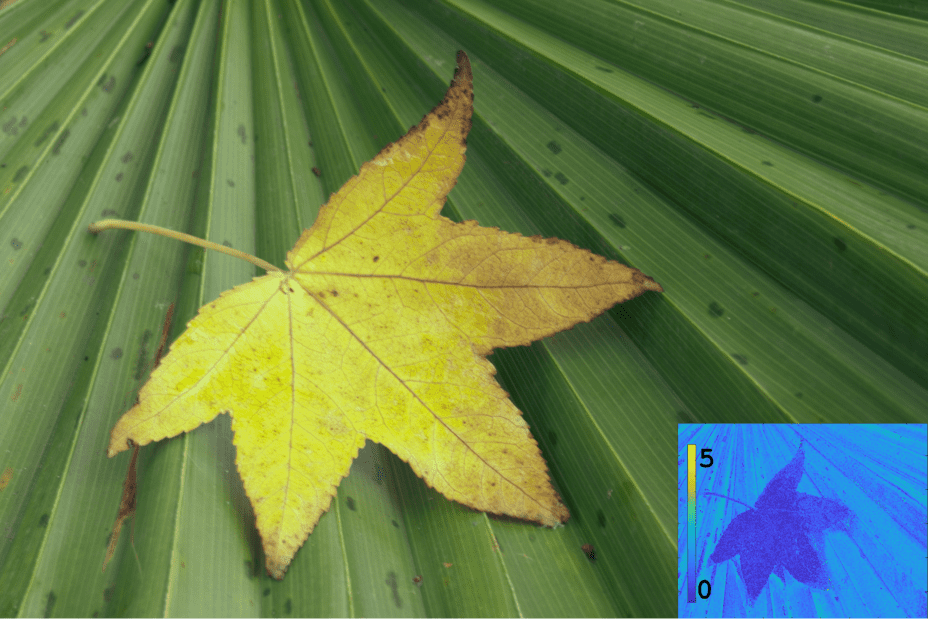}
    \tabularnewline
    \includegraphics[width=0.30\linewidth]{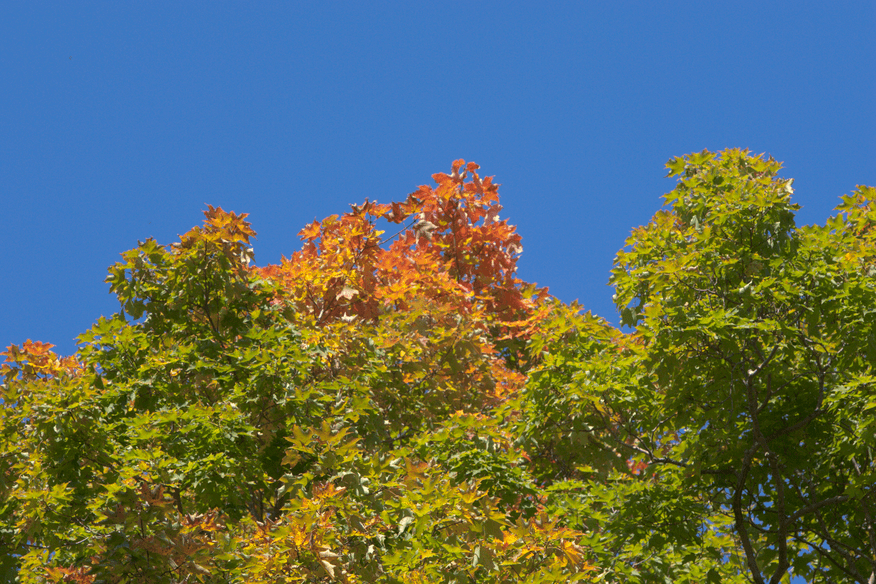} &
    \includegraphics[width=0.30\linewidth]{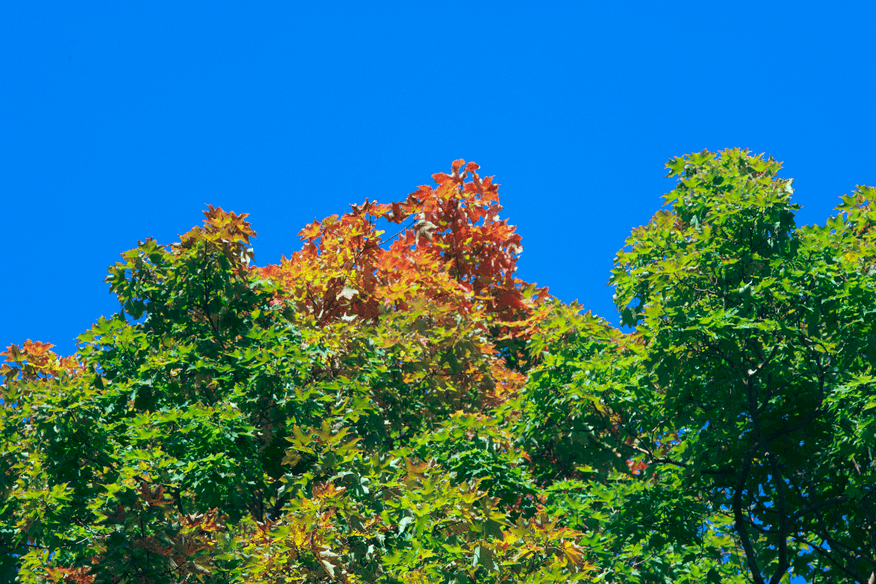} &
    \includegraphics[width=0.30\linewidth]{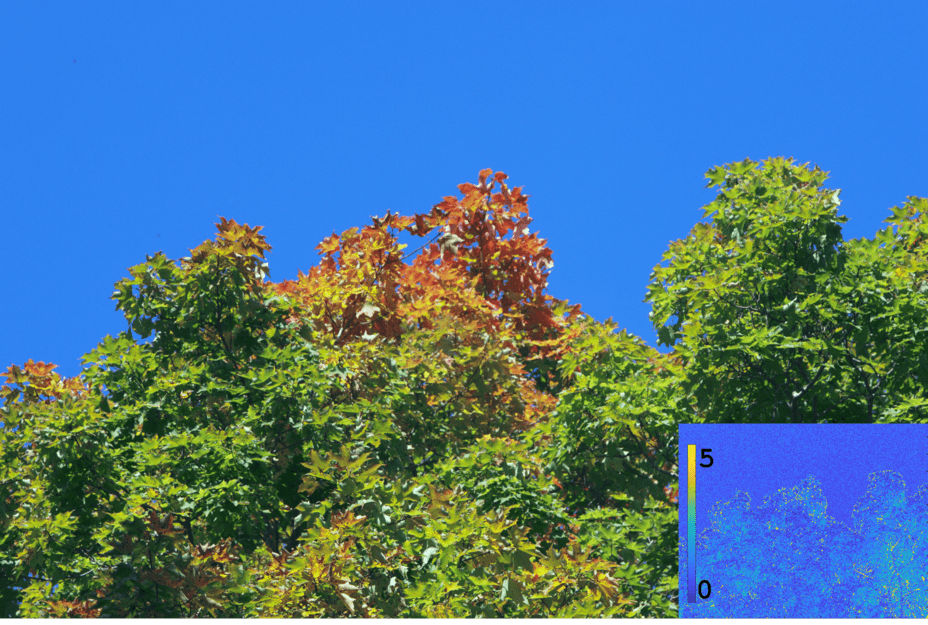}
    \tabularnewline
    \includegraphics[width=0.30\linewidth]{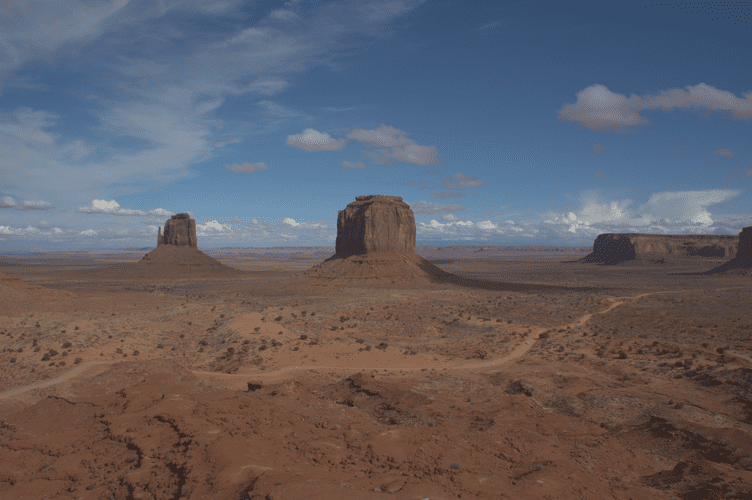} &
    \includegraphics[width=0.30\linewidth]{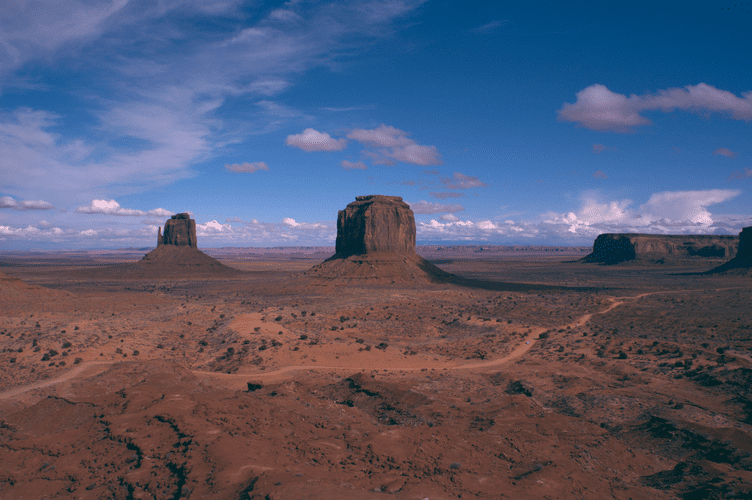} &
    \includegraphics[width=0.30\linewidth]{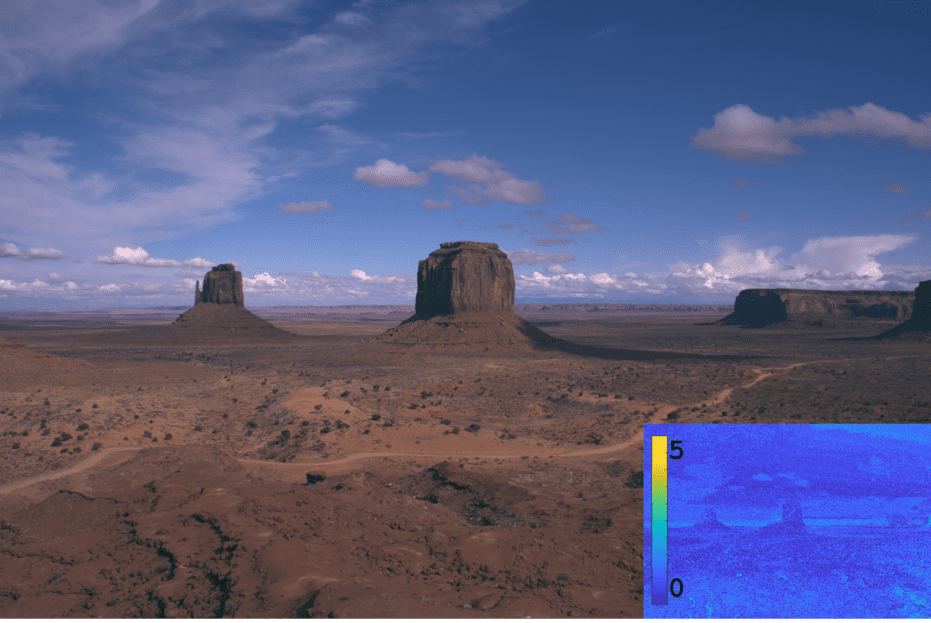} 
    \tabularnewline
    \includegraphics[width=0.30\linewidth]{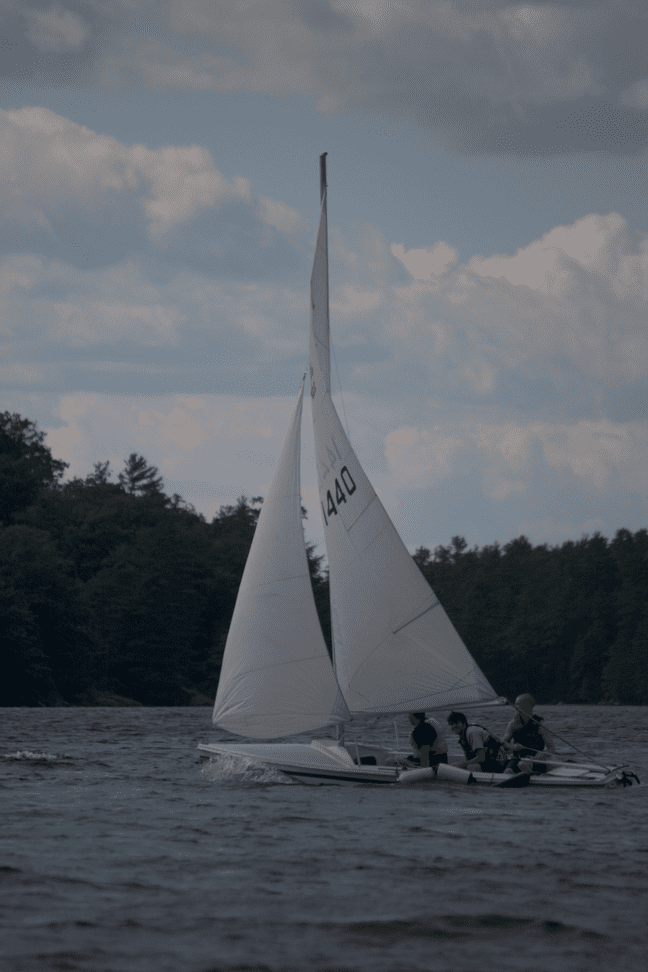} &
    \includegraphics[width=0.30\linewidth]{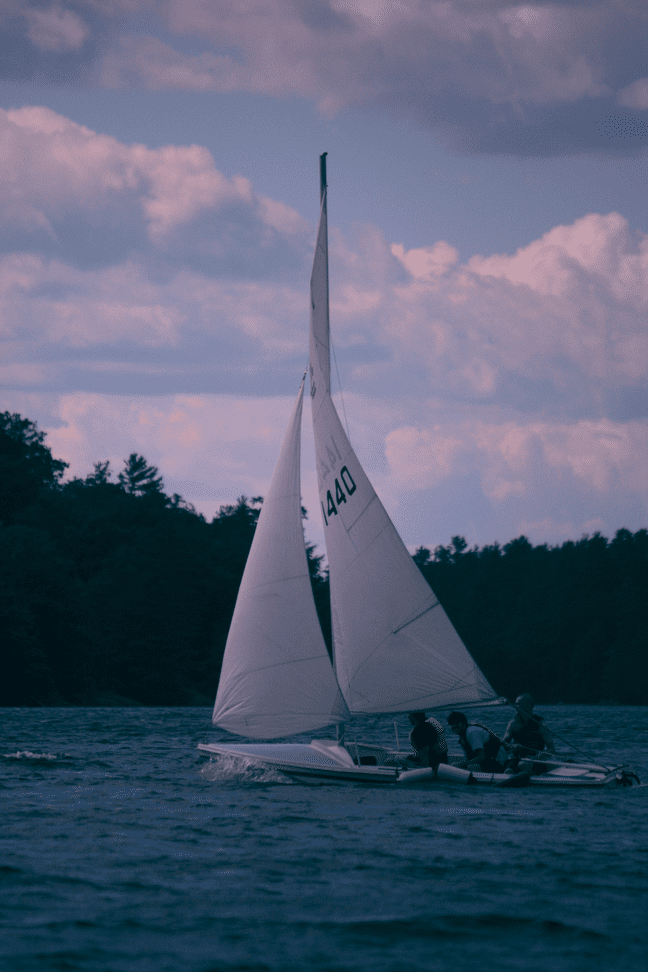} &
    \includegraphics[width=0.30\linewidth]{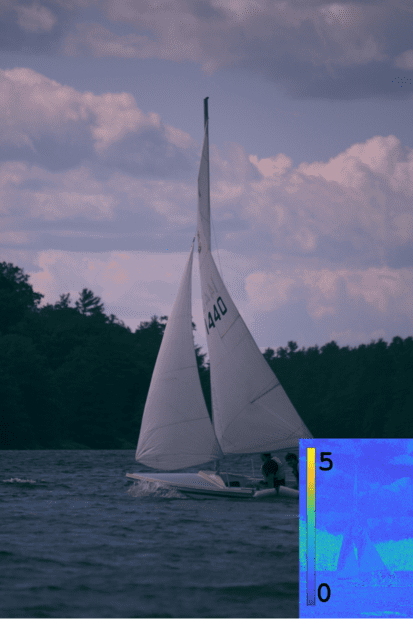}
    \tabularnewline
    Input & Real 3D LUT & Ours NILUT \tabularnewline
    \end{tabular}
    \caption{Results for two different styles (1-2 rows, and 3-4 rows). We can see how the NILUTs are able to reproduce the different styles from professional 3D LUTs. Samples from MIT5K~\cite{fivek}. Note that the NILUT never saw these (or any) real images. Best in electronic version.
    }
    \label{fig:more_results}
\end{figure}

We also present in Fig.~\ref{fig:more_results} results for the case of MLP-Res $N=128$ and $L=2$. We show from left to right, the input image, the ground-truth image processed by real a 3D LUT, and our result. We also display a miniature error map for each of our results. The error map is scaled between 0 and 5 $\Delta$E Units. We provide more results in the supplementary material. 

\begin{figure}[!ht]
    \centering
    \includegraphics[width=0.95\linewidth]{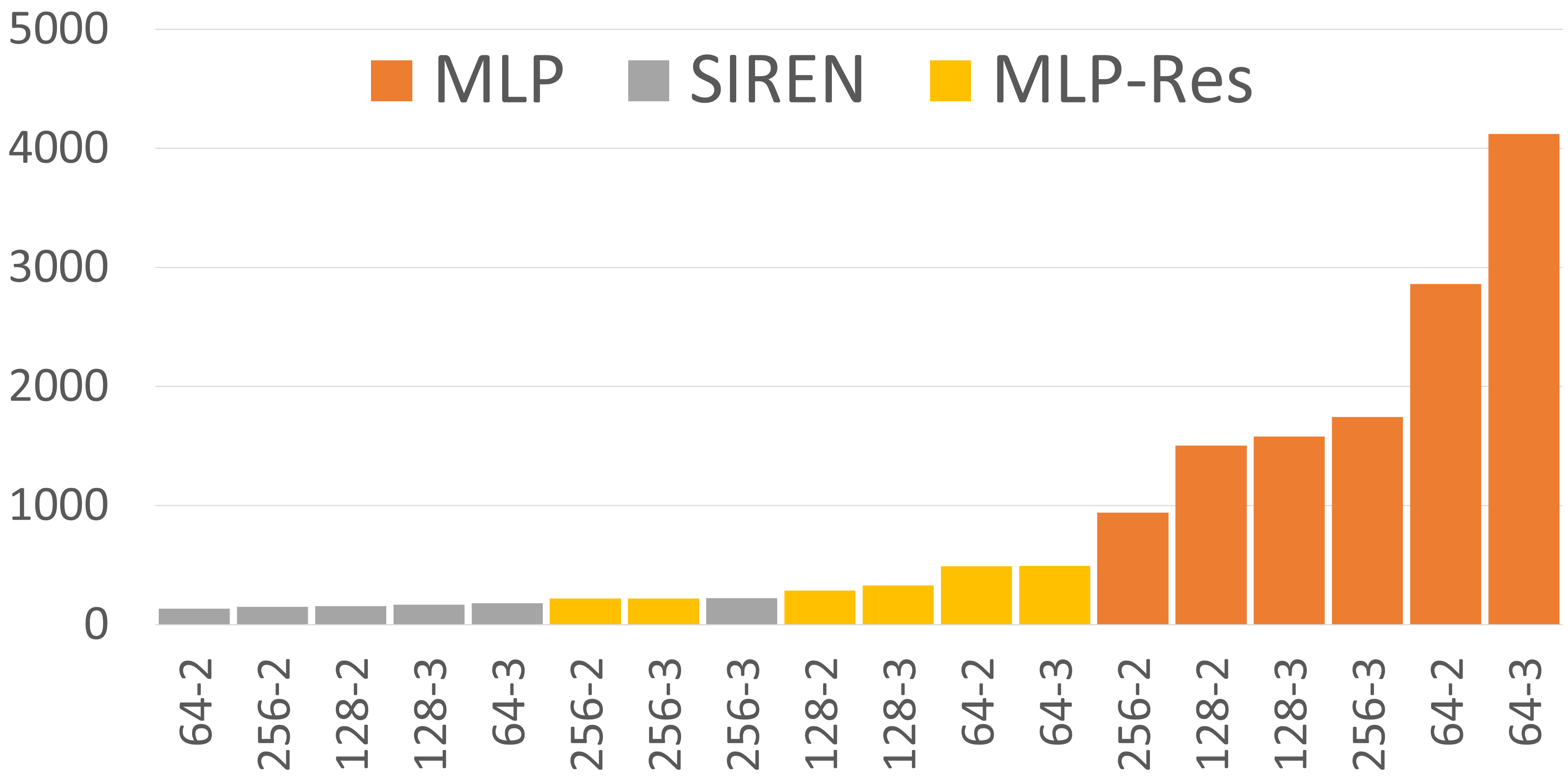}
        \caption{Convergence study. Required number of training steps for each architecture ($N$ neurons, $L$ layers) to achieve over 40dB PSNR at RGB mapping quality -see Fig.~\ref{fig:rgb_map}-.
        \vspace{-2mm}
        }
    \label{fig_iterations}
\end{figure}

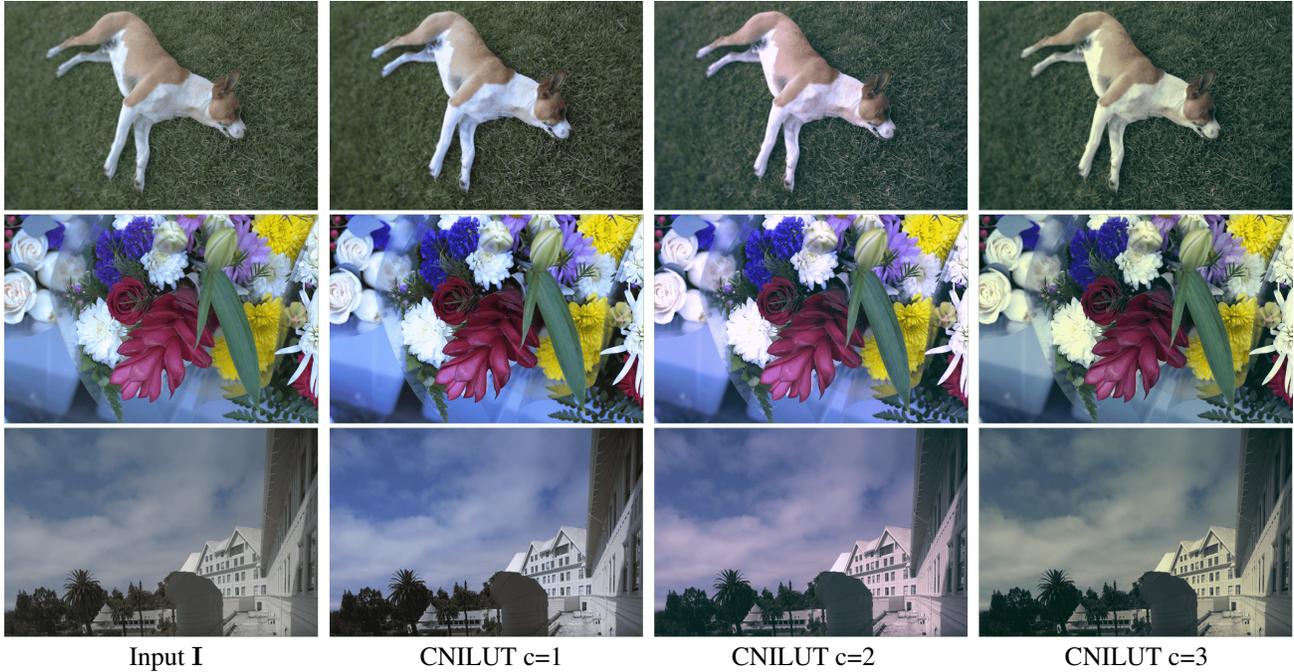
\begin{figure*}[!ht]
    \centering
    \includegraphics[trim={0 4cm 0 0},clip,width=0.98\linewidth]{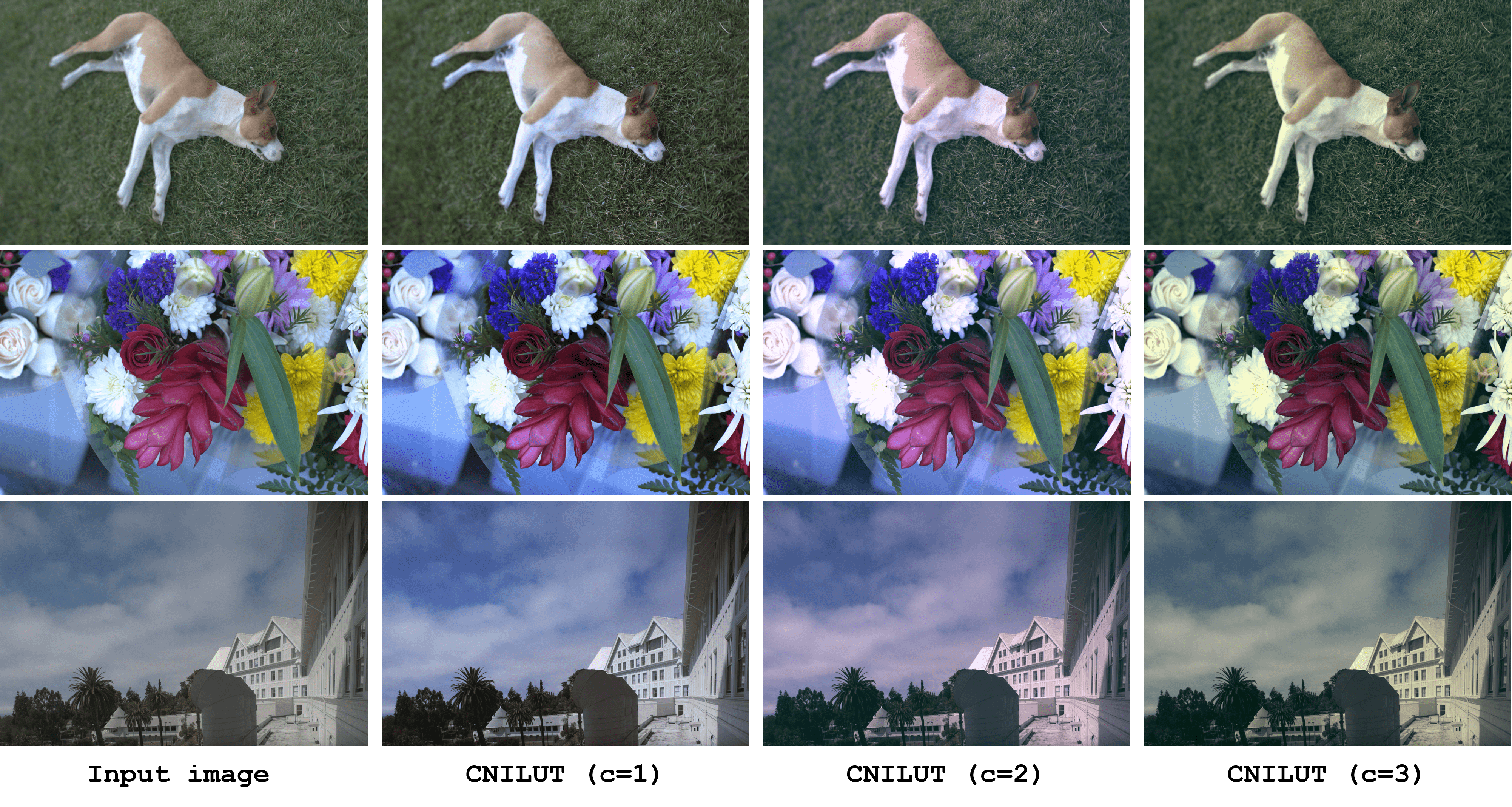}
    \put(-440,-5){Input $\mathbf{I}$}
    \put(-330,-5){CNILUT c=1}
    \put(-212,-5){CNILUT c=2}
    \put(-86,-5) {CNILUT c=3}
    \caption{Results for our conditional NILUTs (CNILUTs). A \textbf{single CNILUT} is able to represent three different styles in a single model. All the images in this figure have less than 3 $\Delta$E error with respect to the real LUT. Best viewed in color.}
    \label{fig:conditional_luts}
\end{figure*}

\vspace{-3mm}
\paragraph{Ablation Studies}
We studied the results for a larger configuration of MLPs under the first evaluation scenario. Results are presented in Table~\ref{tab:ablation}. We can see how for all the cases studied we are able to obtain values that are higher than 40 dB in PSNR and smaller than 1.5 on $\Delta$E, \ie our results have high quality in terms of PSNR and are, on average, indistinguishable from the ground-truth for a human observer. Also, for all the studied architectures we looked at their convergence speed. In Fig.~\ref{fig_iterations} we show for each of the architectures the number of iterations that they required to obtain a result of 40 dB. We can clearly see that the basic MLP needs a larger number of iterations in comparison to SIREN~\cite{sitzmann2020siren} and the Residual-MLP. For the last one, our main architecture MLP-Res, the training is more stable than for SIREN~\cite{sitzmann2020siren}, and we can achieve almost perfect mapping in 4 minutes without using special INR acceleration~\cite{muller2022instant}.

\begin{figure}[!t]
  \centering
   \includegraphics[width=\linewidth]{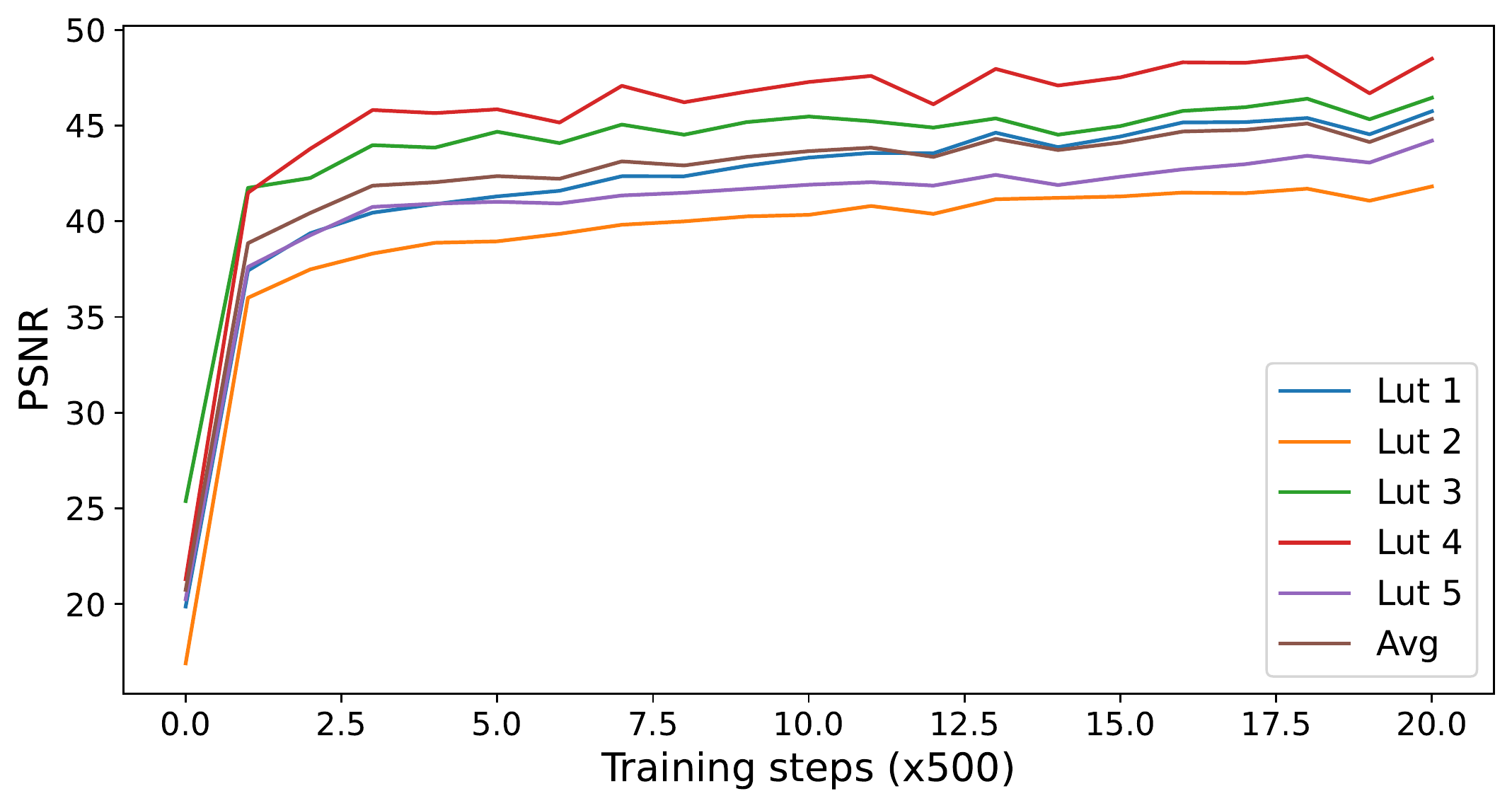}
   \caption{Performance evolution for the CNILUT fitting. Our single CNILUT can accurately learn five different styles from five real 3D LUTs. The reported PSNR is calculated over the five respective reference RGB maps $(\mathcal{M})$. The final average PSNR over the five learned LUTs is 45.34.
   \vspace{-9mm}
   }
   \label{fig:cnilut}
\end{figure}

\begin{table}[t]
    \centering
    \resizebox{\linewidth}{!}{
    \begin{tabular}{c c c c c c}
         \toprule
         Method & N & L & \# Par. (K) & PSNR $\uparrow$ & CIELAB $\Delta$E $\downarrow$ \\
         \midrule
         SIREN~\cite{sitzmann2020siren } & 256 & 2 & 133.3 & 41.95 & 1.45 \\
         SIREN~\cite{sitzmann2020siren } & 256 & 3 & 199.1 & 43.00 & 1.33 \\
         SIREN~\cite{sitzmann2020siren } & 128 & 2 & 33.90 & 44.43 & 1.04 \\
         SIREN~\cite{sitzmann2020siren } & 128 & 3 & 50.40 & 42.74 & 1.43 \\
         SIREN~\cite{sitzmann2020siren } & 64 & 2 & 8.700 & 45.37 & \underline{0.96} \\
         SIREN~\cite{sitzmann2020siren } & 64 & 3 & 12.90 & 43.03 & 1.35 \\
         MLP-Res & 256 & 2 & 133.3 & \underline{45.84} & 1.09 \\
         MLP-Res & 256 & 3 & 199.1 & \textbf{46.19} & \textbf{0.91} \\
         MLP-Res & 128 & 2 & 33.90 & 45.34 & 0.97 \\
         MLP-Res & 128 & 3 & 50.40 & 45.47 & \underline{0.96} \\
         MLP-Res & 64 & 2 & 8.700 & 43.84 & 1.11 \\
         MLP-Res & 64 & 3 & 12.90 & 43.55 & 1.14 \\
         MLP & 256 & 2 & 133.3 & 44.34 & 1.00 \\
         MLP & 256 & 3 & 199.1 & 44.49 & 1.03 \\
         MLP & 128 & 2 & 33.90 & 43.18 & 1.19 \\
         MLP & 128 & 3 & 50.40 & 42.26 & 1.34 \\
         MLP & 64 & 2 & 8.700 & 41.41 & 1.41 \\
         MLP & 64 & 3 & 12.90 & 40.86 & 1.49 \\
         \bottomrule
    \end{tabular}}
    \caption{Ablation study of different MLPs. Results are computed as the differences between the NILUT generated map $\Phi (\mathcal{M})$ and the real 3D LUT map $\phi (\mathcal{M})$. Metrics are the average over the five 3D LUTs in our dataset.
    }
    \vspace{-3mm}
    \label{tab:ablation}
\end{table}

\begin{figure*}[!ht]
  \centering
  \setlength\tabcolsep{1.5pt}
  \begin{tabular}{cccc}
    \includegraphics[width=0.24\linewidth]{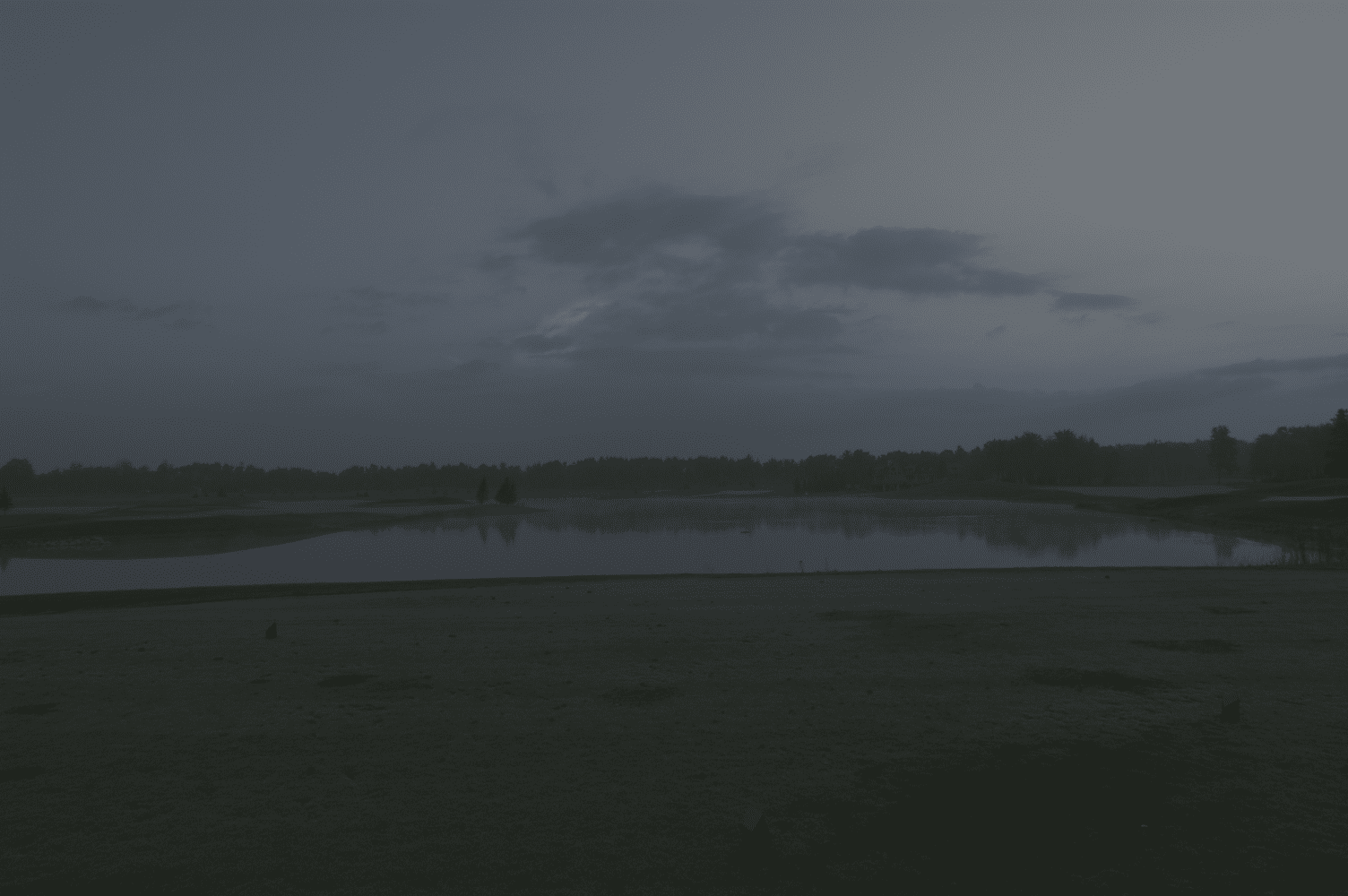} &
    \includegraphics[width=0.24\linewidth]{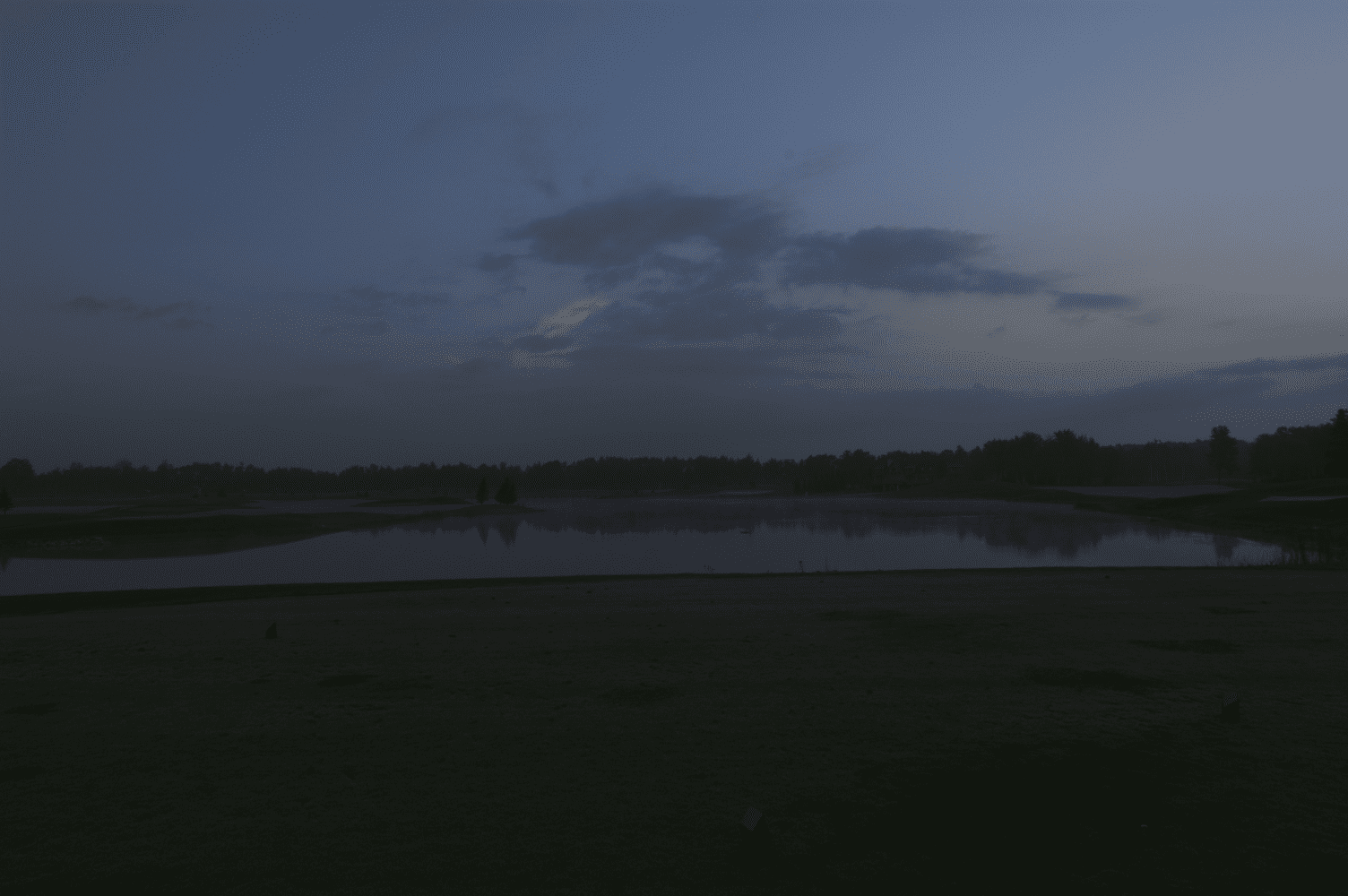} &
    \includegraphics[width=0.24\linewidth]{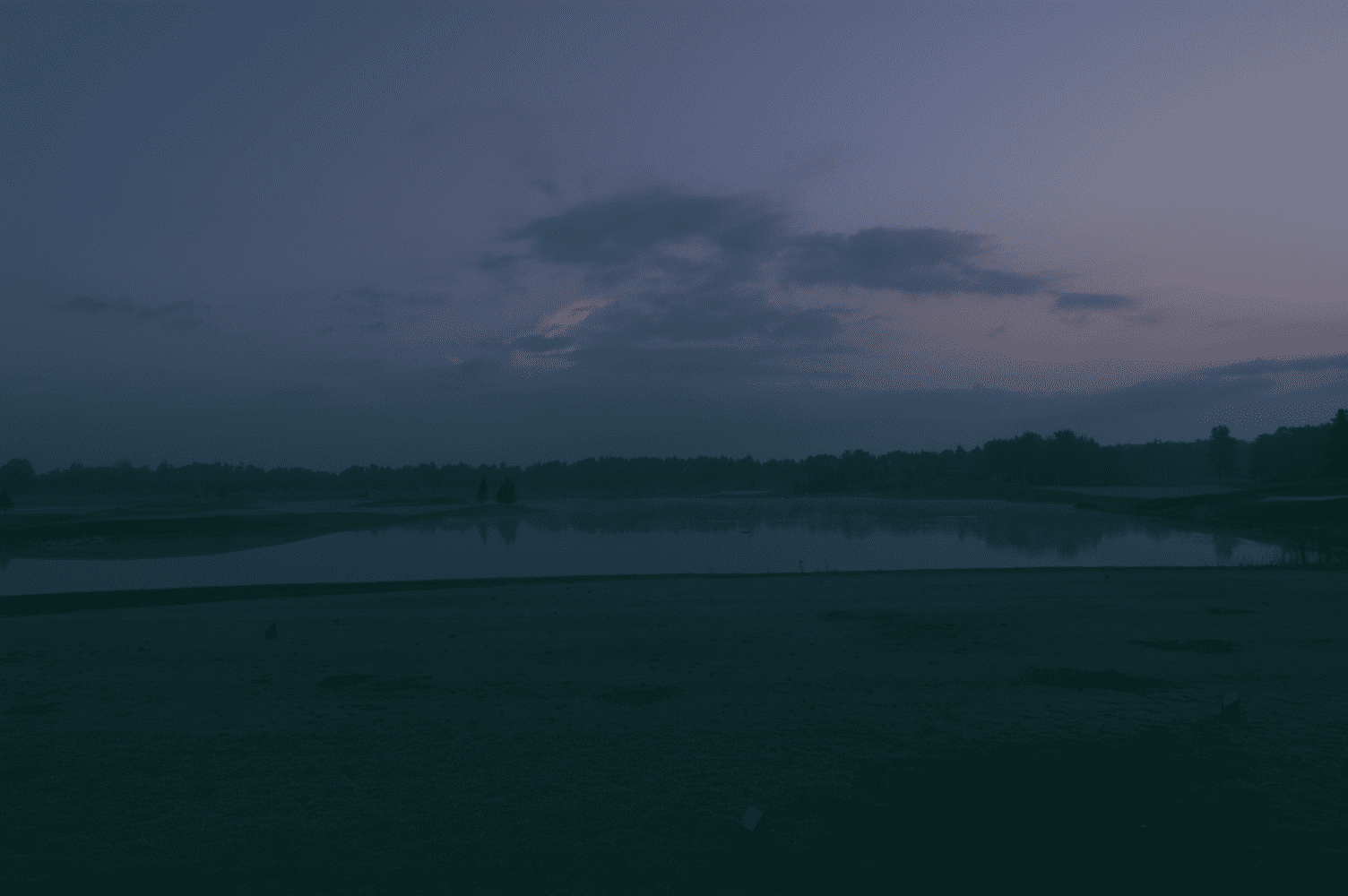} &
    \includegraphics[width=0.24\linewidth]{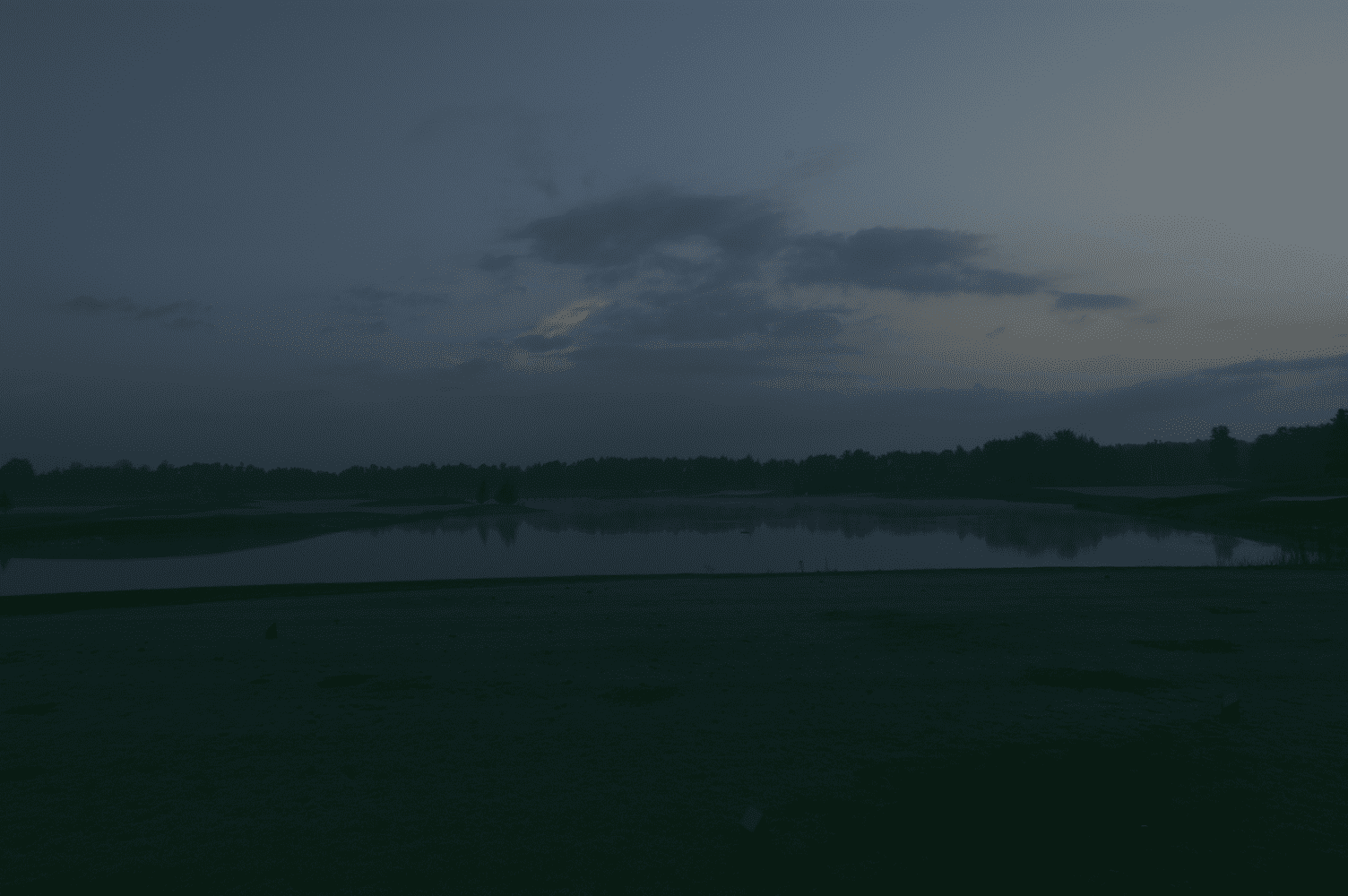} \tabularnewline
    Input & CNILUT $[1,0,0]$ & CNILUT $[0,1,0]$ & CNILUT $[0.5,0,0.5]$
    \tabularnewline
    \includegraphics[width=0.24\linewidth]{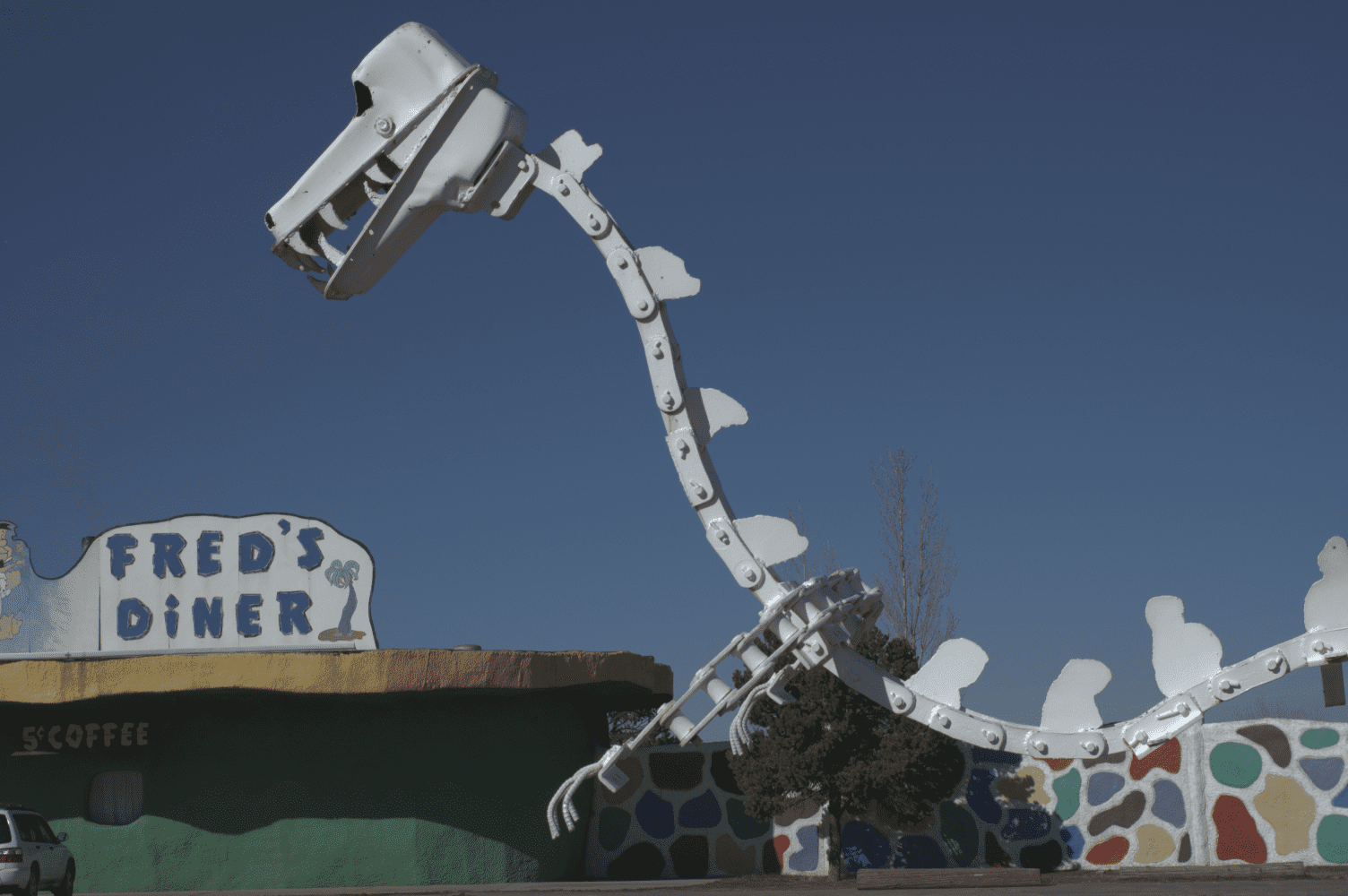} &
    \includegraphics[width=0.24\linewidth]{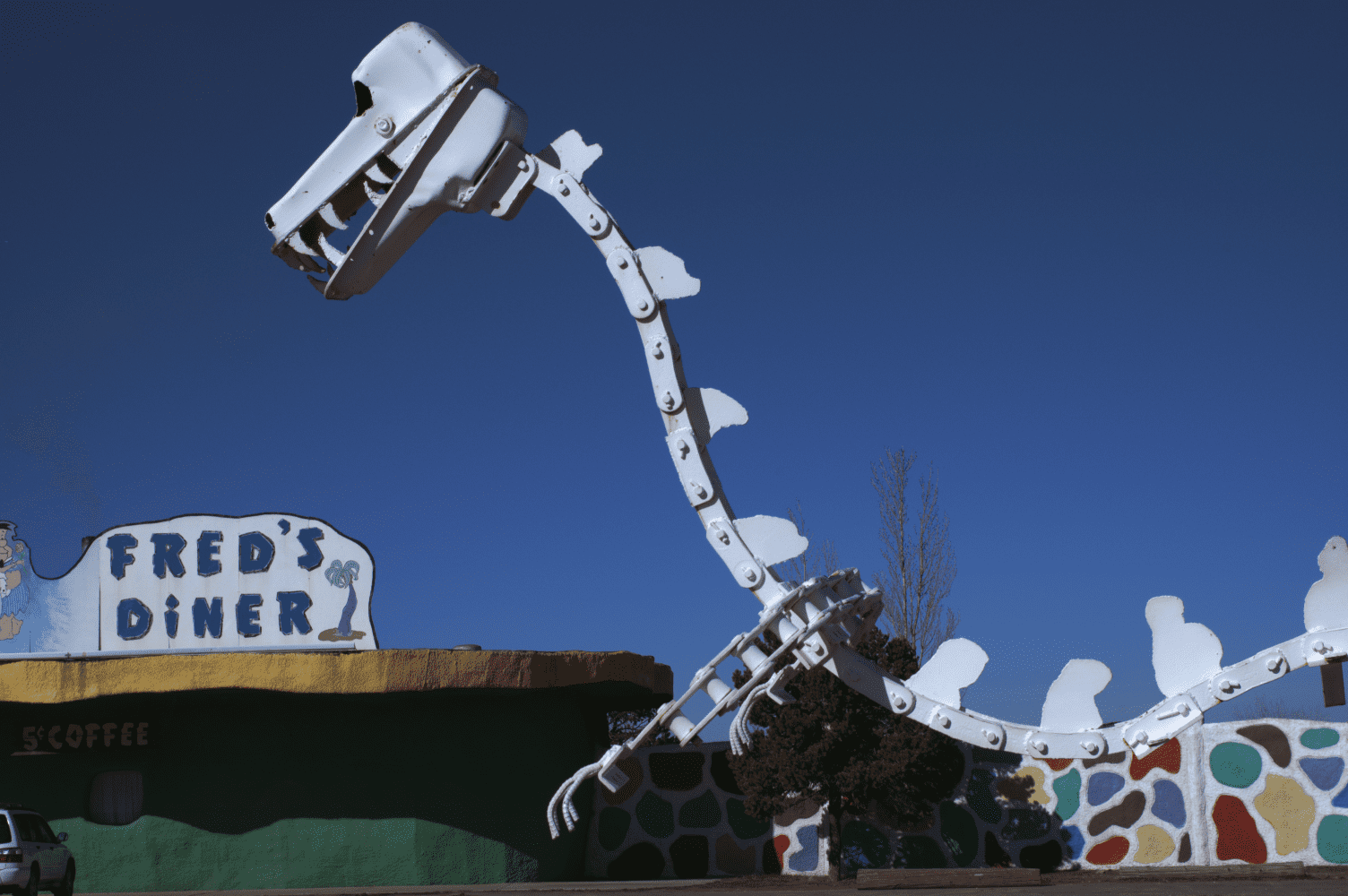} &
    \includegraphics[width=0.24\linewidth]{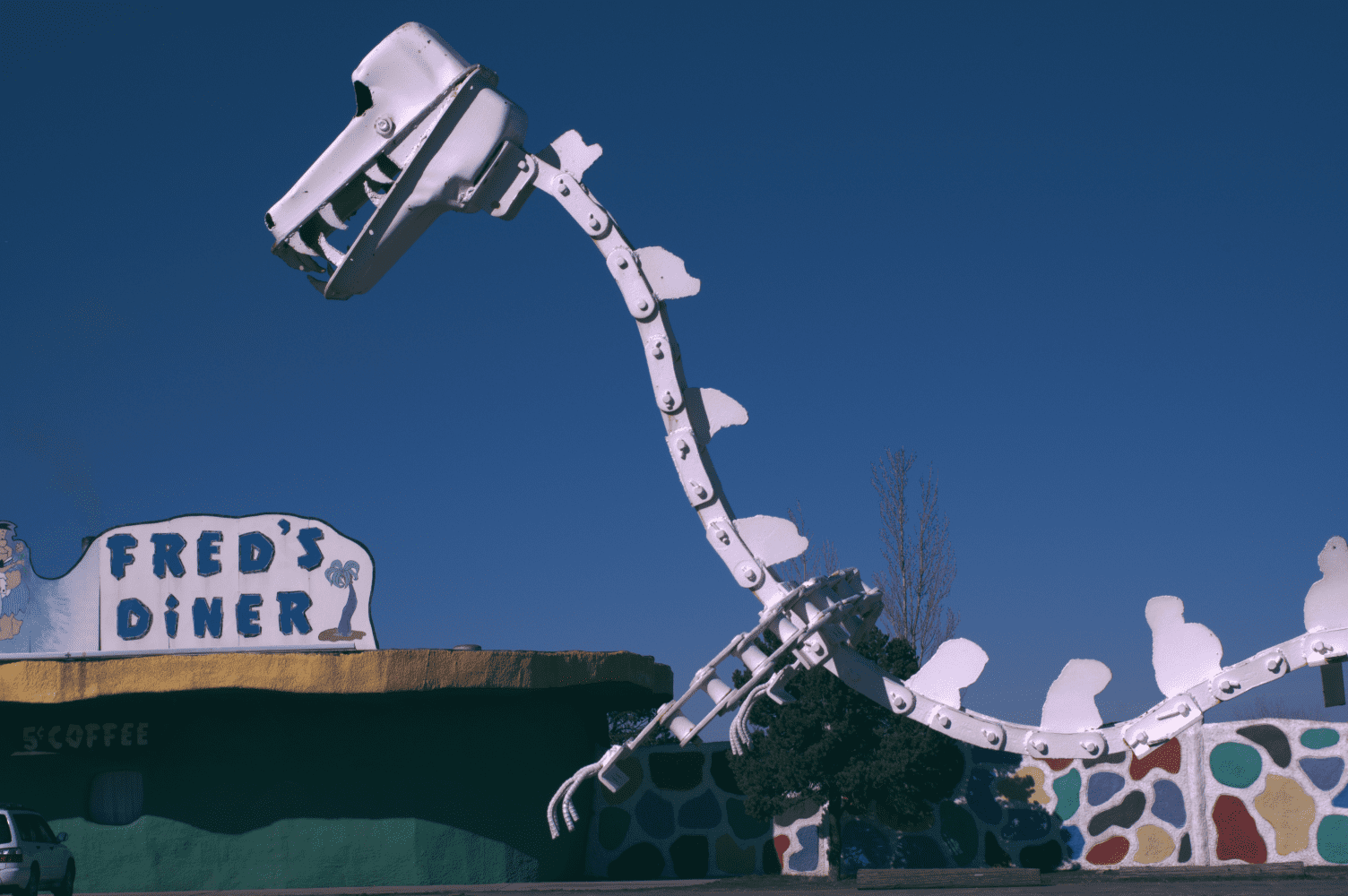} &
    \includegraphics[width=0.24\linewidth]{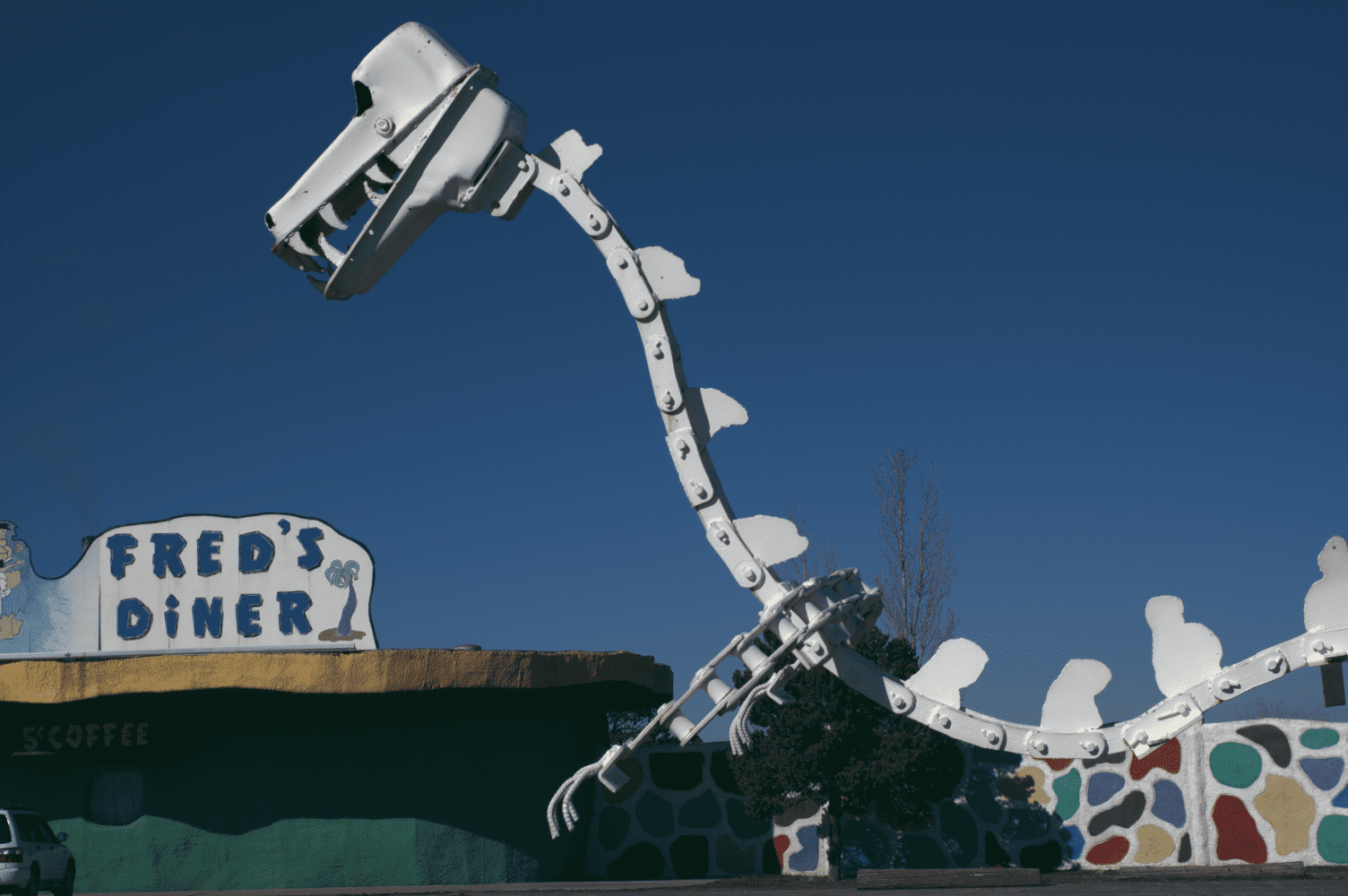} \tabularnewline
    Input & CNILUT $[1,0,0]$ & CNILUT $[0,0,1]$ & CNILUT $[0.25,0.5,0.25]$
    \tabularnewline
  \end{tabular}
   \caption{Example of the blending capabilities of the proposed CNILUT. In particular, the proposed MLP-Res 128x2 is trained on a set of three 3D LUTs, which achieves an average RGB mapping quality of 43.72dB. We show the three learned styles (one-hot encoded condition vectors), and two random ``blendings" given the specified condition vectors.}
   \label{fig:blending}
\end{figure*}

\subsection{Conditional NILUT}
\label{ssc:cnilut}

As we introduced in Section~\ref{sec:ours}, our NILUT can be conditioned to different styles (CNILUT), and by doing this, we can learn implicitly multiple 3D LUT transformations using a single NILUT. This novel feature would allow reducing the memory requirements of storing and calling multiple 3D LUTs in a camera pipeline~\cite{karaimer2016software}. In our experiments, we set to three and five the number of 3D LUTs to learn using this approach. This however denotes a clear \emph{trade-off} since, in exchange of this ability, CNILUTS require longer and more complex training, and there is a slight performance degradation in comparison to learning three/five separated -and specialized- NILUTs. 
This said we are able to obtain consistent values larger than $42$ dBs in PSNR and errors smaller than $1.5$ $\Delta$E in the RGB mapping, therefore being almost unnoticeable for human observers. Fig.~\ref{fig:cnilut} shows the evolution of the training of the CNILUT using an MLP-Res 256x2. Note that as before, we train using five RGB maps, and use as the main metric -in this case- the average \texttt{avg} over the five representations. As we show, some LUTs are easier to learn than others, yet we can learn five LUTs with an average RGB mapping PSNR of 45.34dB, and $\Delta$E 0.85.
In Table~\ref{tab:cnilut}, we provide an ablation study where we show the performance of different MLP-Res architectures when learning 3 and 5 different 3D LUTs. The error is computed as the average for the different 3D LUTs. Given the large PSNR and errors smaller than $1.2$ $\Delta$E in the RGB mapping, we can confirm that the difference between the CNILUT mapping and the corresponding three/five 3D LUTs is almost unnoticeable for human observers~\cite{sharma2017digital}.
The main difference between CNILUT and NILUT is the training: CNILUT requires longer training to obtain good performance. We provide this study in the supplementary material. 

\vspace{-5mm}
\paragraph{Multi-Styles Qualitative Results}
In Fig.~\ref{fig:conditional_luts} we present some qualitative results of our CNILUT. 
A single CNILUT can successfully emulate the behavior of three different 3D LUTs by imposing the corresponding condition vectors. 

\vspace{-4mm}
\paragraph{Memory Efficiency}
From previous experiments, we can conclude that a CNILUT can compress the representation of five 3D LUTs, without additional computational cost \ie the number of operations due to the integration of the condition vector does not affect the runtime.
Considering a standard 33-dimensional 3D LUTs~\cite{zeng2020lut, yang2022adaint} with $\approx107K$ parameters, if stored as FP32 would require $\approx0.43$MB. Our NILUT has 8.7K parameters ($\approx0.032$MB) and can accurately reproduce the behavior and properties of such complex 3D LUTs, which implies a compression of $13\times$. This facilitates its storage in mobile devices with limited on-chip memory.

\begin{table}[]
    \centering
    \begin{tabular}{c c c c c}
         \toprule
         \multirow{2}{*}{Method} & \multicolumn{2}{c}{Learn \textbf{x3}} & \multicolumn{2}{c}{Learn \textbf{x5}} \\
         & PSNR~$\uparrow$ & $\Delta$E~$\downarrow$ & PSNR~$\uparrow$ & $\Delta$E~$\downarrow$ \\
         \midrule
         128x2 & 43.72 & 1.07 & 42.67 & 1.19  \\
         128x3 & 44.03 & 0.99 & 43.71 & 1.01  \\
         256x2 & \underline{45.37} & \underline{0.90} & \underline{45.34} & \underline{0.85}  \\
         256x3 & \textbf{47.26} & \textbf{0.74} & \textbf{46.05} & \textbf{0.8}  \\
         \bottomrule
    \end{tabular}
    \caption{Conditional NILUT ablation study. All the reported architectures are MLP-Res. We report results when learning 3 and 5 different 3D LUTs styles in a single CNILUT. We train our CNILUT for 10000 steps.
    \vspace{-2mm}
    }
    \label{tab:cnilut}
\end{table}

\vspace{-4mm}
\paragraph{Blending Styles}
\label{sec:blending}
The derivation of our CNILUT also allows us to blend among different styles by just modifying the values of the condition vector. When this vector is a one-hot encoder we have one of the basis implicit 3D LUTs; but we can generalize this vector as a set of blending weights. 
%
The multi-style blending occurs as an implicit interpolation within the neural network itself. This is the same principle as in SIREN~\cite{sitzmann2020siren} (interpolation of pixels for 2D images) and NeRF~\cite{mildenhall2021nerf}. 
We analyze the output of blending the test images with weights $[0.33, 0.33, 0.33]$, and compare with the linear interpolation of the real processed 2D RGB images, the PSNR is 40.05dB, indicating the implicit CNILUT blending is equivalent to explicit images interpolation.

We provide examples of our blending capabilities in Fig.~\ref{fig:blending}. We believe the blending property itself represents an interesting future work. We provide more details in the supplementary material.

\begin{figure}[!ht]
  \centering
  \resizebox{\linewidth}{!}{
  \begin{tabular}{ccc}
    \includegraphics[width=\linewidth]{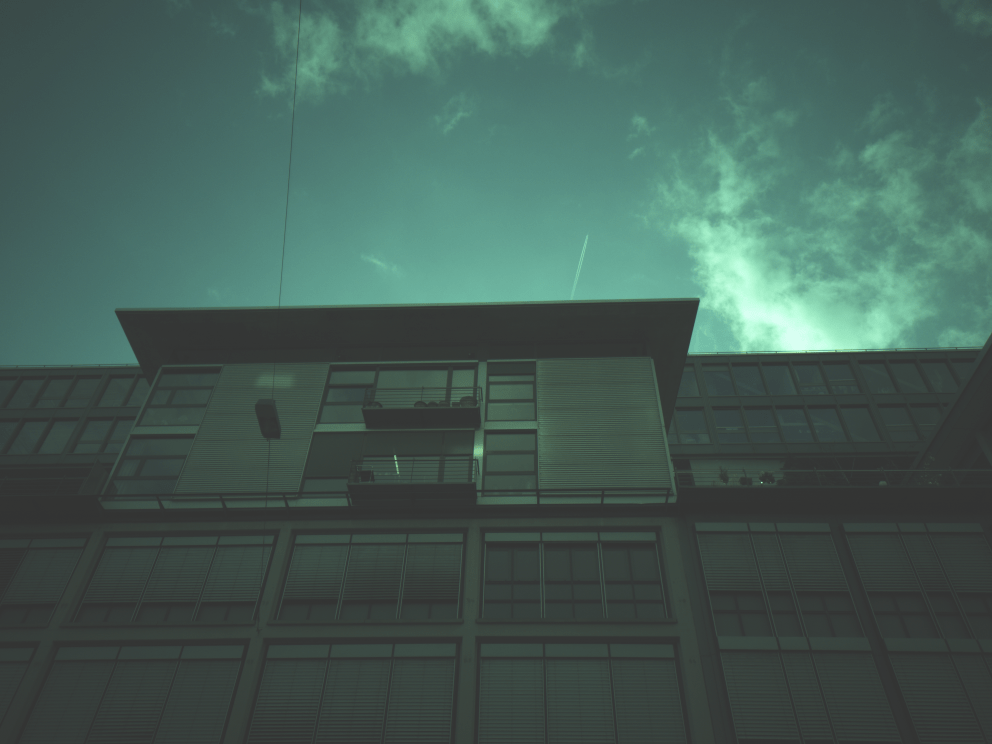} &
    \includegraphics[width=\linewidth]{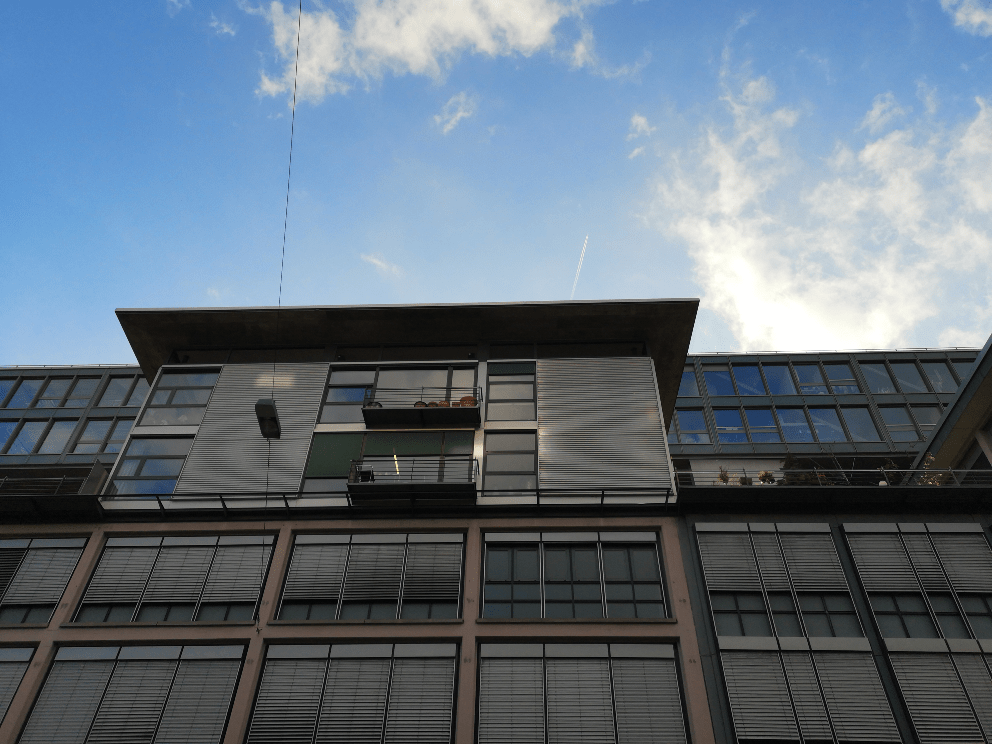} &
    \includegraphics[width=\linewidth]{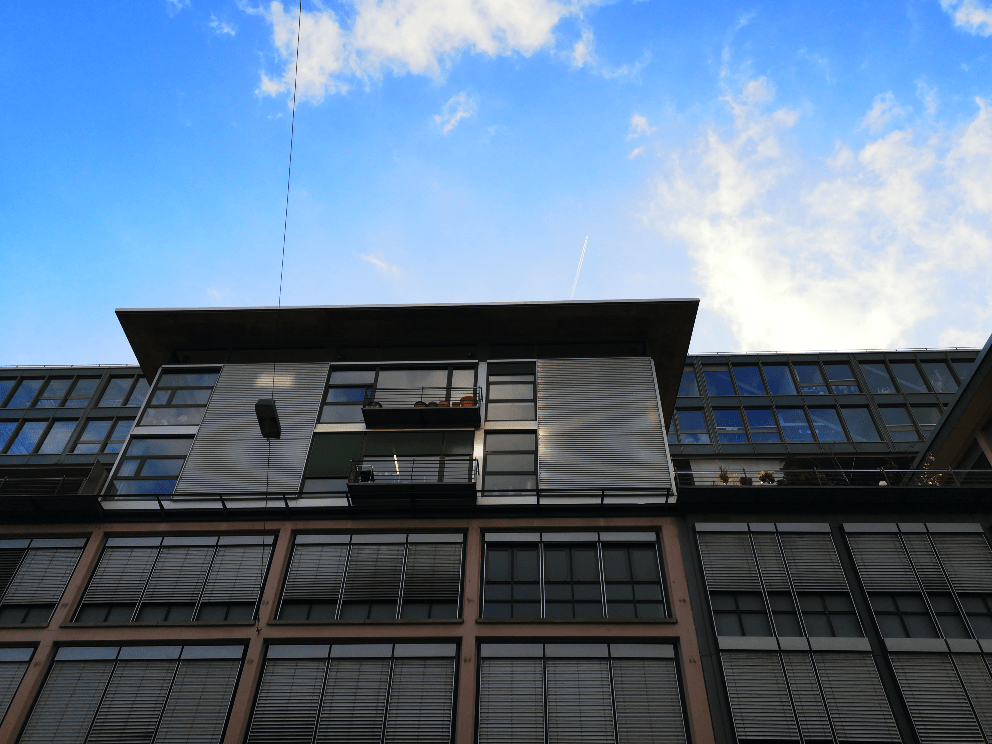} \tabularnewline
    \tabularnewline
    \includegraphics[width=\linewidth]{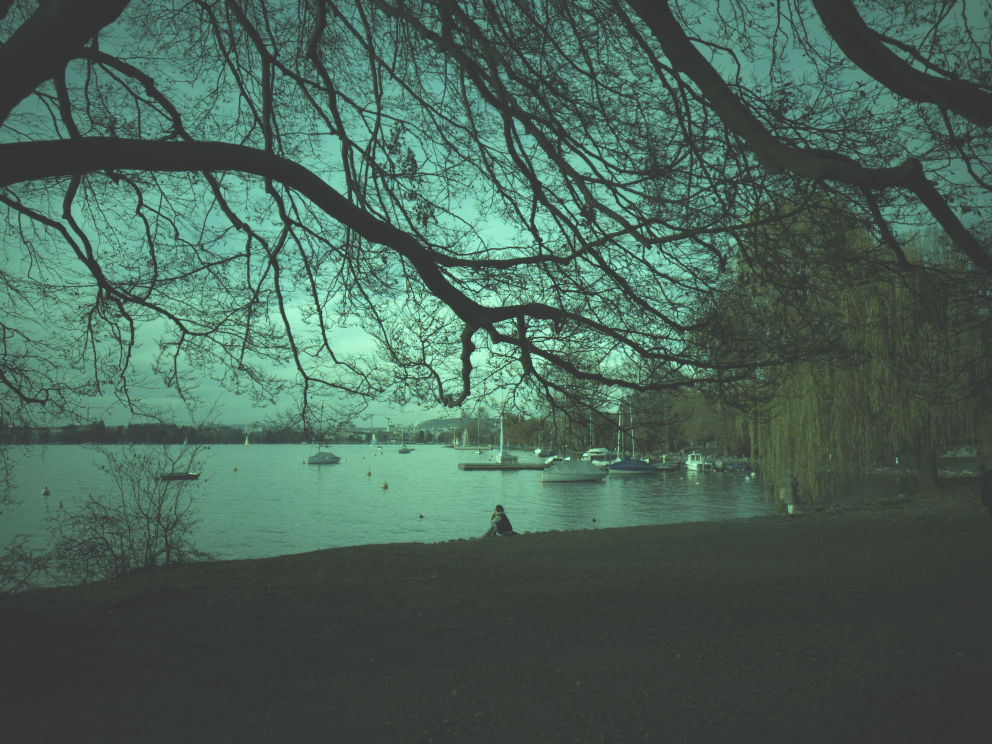} &
    \includegraphics[width=\linewidth]{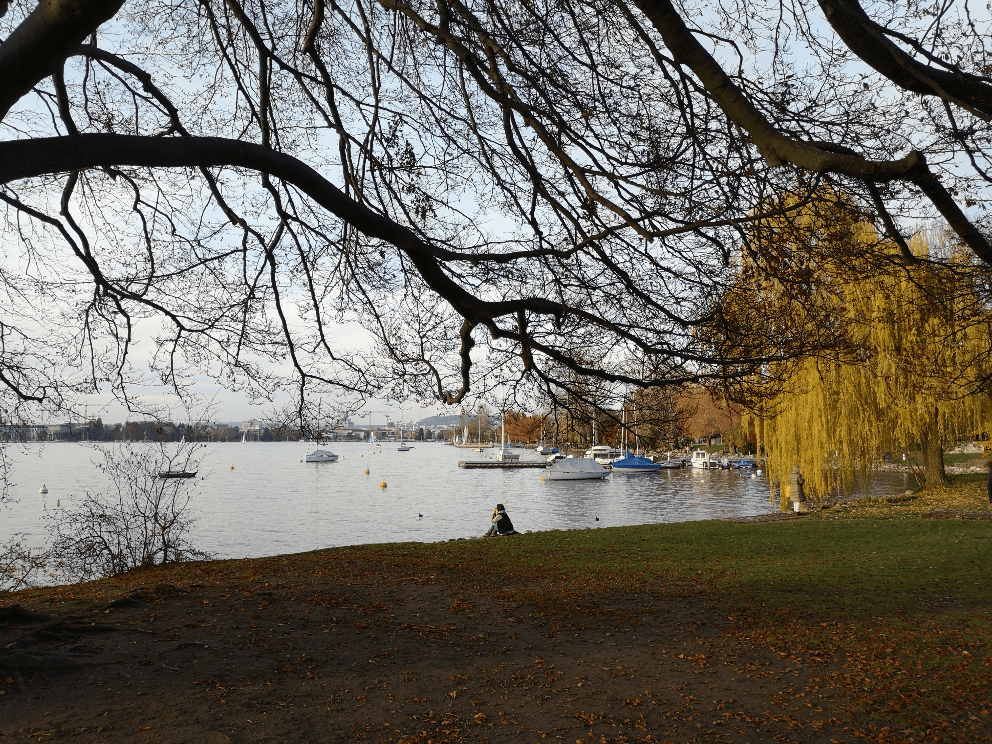} &
    \includegraphics[width=\linewidth]{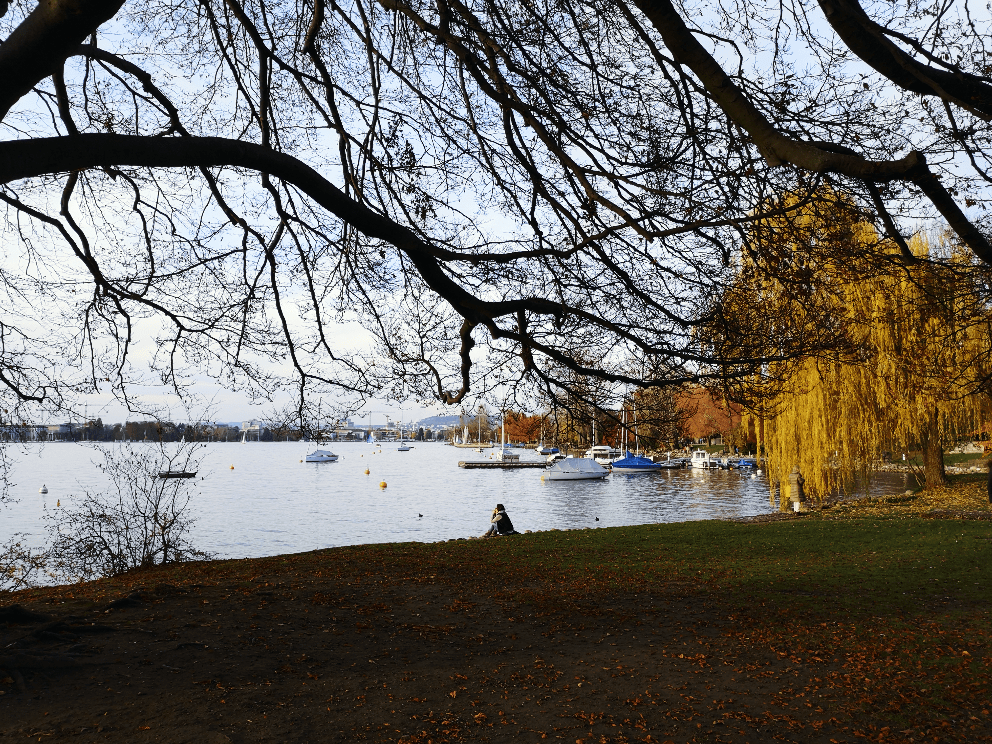} \tabularnewline
  \end{tabular}}
   \caption{NILUT as plug-in module to further enhance learned ISPs. (Left) \textbf{RAW} image after demosaicing. (Middle) \textbf{RGB} image produced using a learned ISP designed for NPU~\cite{ignatov2021realnpu, ignatov2018ai}. (Right) Enhanced image using our \textbf{NILUT}.
   \vspace{-2mm}
   }
   \label{fig:isp-lut}
\end{figure}

\subsection{Plug-in deep learning based ISPs}
\label{sec:nilut-isp}

We explore the integration of NILUTs into modern learned ISPs~\cite{conde2022model, tseng2022neural}. 
Our NILUT is a possible plug-and-play module to further enhance the colors and apply different styles. Moreover, it is differentiable, which facilitates end-to-end ISP optimization.
In Fig.~\ref{fig:isp-lut}, we show how to complement a learned ISP~\cite{ignatov2021realnpu} designed to run -and tested- on smartphones, using our NILUTs to enhance the image further and manipulate colors. 

This is a core step in any traditional ISP~\cite{karaimer2016software}. Further work on training or enhancing ISPs is out of the scope of this work, as it is a research topic by itself~\cite{liang2021cameranet, liu2022deep, ignatov2020replacing, tseng2022neural}.

\subsection{Applications}
\label{sec:app}

\begin{figure}[t!]
    \centering
    \includegraphics[width=0.85\linewidth]{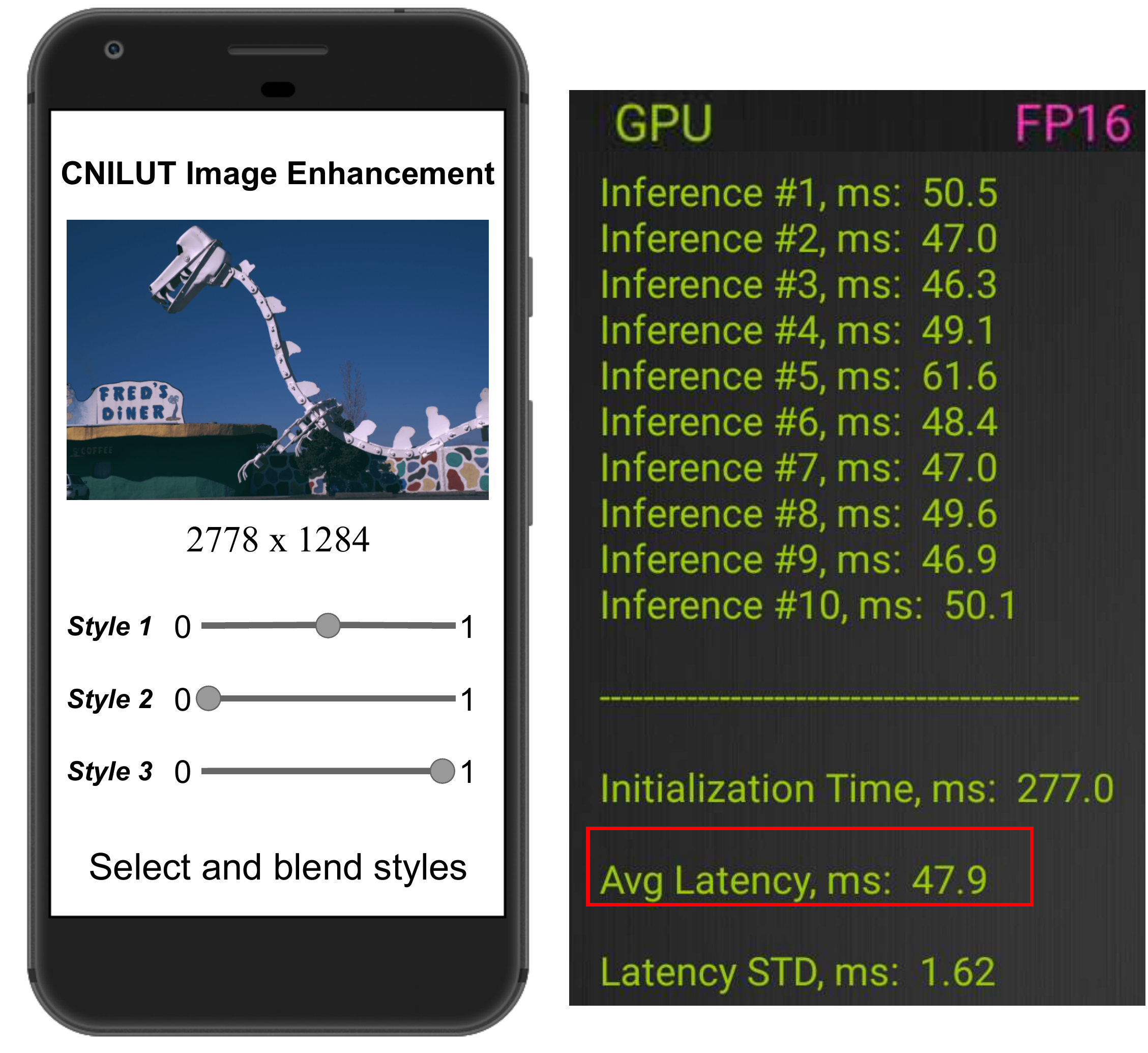}
    \caption{(L) Sample image enhancement application on mobile devices based on the proposed CNILUT. (R) Deployment using~\cite{ignatov2018ai}. See also Section~\ref{sec:blending} and Figure~\ref{fig:blending}.}
    \vspace{-1.5mm}
    \label{fig:app}
\end{figure}

We provide a demo application in Fig.~\ref{fig:app}. Following Section~\ref{sec:nilut-isp}, we deploy a small CNILUT (32x2) on mobile devices using \textit{AI Benchmark}~\cite{ignatov2018ai} and provide the results in Tab.~\ref{tab:run}. The model can process 4K images in smartphones without memory problems at $\approx10$ FPS on regular GPUs (2020). 
The CNILUT (3 styles) is just 4KB in comparison to the 1.2 MB ($3 \times 0.4$MB) of the three complete 3D LUTs.

This technique allows controllable color manipulation by just changing the input condition vectors. By definition, CNILUT is a pixel-wise transformation, therefore the structure of the image is perfectly preserved.

\subsection{Limitations and Other Methods}
\label{sec:limitations}

Despite the promising results of learning 3D LUTs as INRs, there are some limitations that we must consider.

Firstly, as we previously discussed in Section~\ref{sec:discussion}, we understand that a neural network (\eg NILUT) is limited and cannot surpass the efficiency of a lookup operation.

Secondly, by design and mathematical convenience, other methods for learning 3D LUTs such as Zeng~\etal~\cite{zeng2020lut} also achieve ``perfect" mapping (\ie $>50$dB). However, despite the great fitting, the application of classical 3D LUTs on mobile devices is not trivial as discussed in Sec.~\ref{sec:discussion}. 

Thirdly, many previous approaches~\cite{zeng2020lut, yang2022adaint, Yang_ECCV} did not focus on learning explicit 3D LUTs, but instead focused on constructing a mapping between RGB and enhanced RGB (\eg ``Expert C" in MIT5K~\cite{fivek}), which was not necessarily limited to 3D LUT operations.
We aim to complement these approaches~\cite{zeng2020lut, yang2022adaint, Yang_ECCV, wang2021real, zhao2022learning} with the proposed NILUT that offers multi-style functionality and is suitable for modern mobile devices.

\begin{table}[t!]
    \centering
    \footnotesize
    \begin{tabular}{c c c}
    \toprule
    Input resolution    & Mali-G77 MC9 (ms)~$\downarrow$ & Adreno 620 (ms)~$\downarrow$ \\
    \midrule
    $1920 \times 1080 $ & $47.9 \pm 1.62$   & $75.0 \pm 3.28$ \\
    $2778 \times 1284 $ & $73.9 \pm 1.62$   & $135.0 \pm 3.63$ \\
    $3840 \times 2160 $ & $121.0 \pm 1.41$  & $286.0 \pm 2.92$ \\
    \bottomrule
    \end{tabular}
    \vspace{0.5mm}
    \caption{
    CNILUT deployment on two mid-level smartphone GPUs. 
    We report the average image processing runtime over 10 runs $\pm$ the std. deviation (see Figure~\ref{fig:app}).
    }
    \vspace{-4.5mm}
    \label{tab:run}
\end{table}

\section{Concluding remarks}

We have introduced NILUTs, a new approach for modeling 3D LUTs as INRs. Our NILUTs are implicitly defined, continuous, 3D color transformations parameterized by a neural network. They present several advantages: i) are memory-efficient and can run on mobile devices, therefore, are easy to integrate into modern deep learning ISPs; ii) can perform multiple style modifications with a single architecture; and iii) can compute blend between different styles. 
The novel multi-style blending formulation allows controllable image enhancement and customization.
Quantitative and qualitative results demonstrate the effectiveness of our newly defined NILUTs for image enhancement. We have also curated a dataset of 3D LUTs and images for evaluation of color manipulation methods.


\section*{Acknowledgements}

This work was partially supported by the Alexander von Humboldt Foundation. 

This study was funded in part by the Canada First Research Excellence Fund for the Vision: Science to Applications (VISTA) program, an NSERC Discovery Grant, and an Adobe Gift Award. 

JVC was supported by Grant PID2021-128178OB-I00 funded by MCIN/AEI/10.13039/501100011033, ERDF "A way of making Europe", the Departament de Recerca i Universitats from Generalitat de Catalunya with reference 2021SGR01499, and the "Ayudas para la recualificación del sistema universitario español" financed by the European Union-NextGenerationEU. 

\appendix

\section{Implementation Details}

We develop the models using PyTorch framework and two NVIDIA RTX 3090. The MLP networks are designed based on SIREN (and variants)~\cite{sitzmann2020siren}. The models are trained using fixed learning rate $1e^{-3}$ and Adam optimizer~\cite{kingma2014adam} until convergence (\eg{ 5000 steps, $\sim 4$ minutes}). Note that as we show in Fig.~\ref{fig_iterations}, convergence depends on the architecture. For the  experiments in Tables~\ref{tab:ablation2}, \ref{tab:ablation},  \ref{tab:cnilut} we do not use RGB maps of dimension $4096 \times 4096 \times 3$  (\ie{ complete 16M values}), instead, we use a reduced map of dimension $2048 \times 1024 \times 3$ which contains one of each two possible values in the R,G, and B channels (\ie{ $128^3$ values}), which requires less memory and allows faster experimentation.

For training the conditional NILUT we use three/five different 3D LUTs, at each step we feed the three/five condition vector and RGB map into the network and accumulate the three/five different loss terms (one for each learned LUT). We concatenate the map and condition vector to obtain an input with dimension $[h\times w,6]$/$[h\times w,8]$. Since we learn three/five different 3D LUTs using a single CNILUT, we need to train at least 4000 steps to obtain reasonable results. Once the CNILUT is trained, we can further fine-tune it to perform blending by yielding random condition vector weights (\ie{ softmax weights}) and the corresponding blended outputs; these represent plausible convex combinations of the three basis 3D LUTs. 

{\small
\bibliographystyle{ieee_fullname}
\bibliography{egbib}
}

\end{document}